\documentclass[letterpaper,11pt]{article} 

\usepackage{amsmath}
\usepackage{amssymb}
\usepackage[comma,longnamesfirst]{natbib}
\usepackage[dvips]{epsfig}
\usepackage{dcolumn}
\usepackage{enumerate}
\usepackage{hhline}
\usepackage{dsfont}
\usepackage{afterpage}
\usepackage{arydshln}
\usepackage{graphicx}
\usepackage{color}
\usepackage[usenames,dvipsnames]{xcolor}
\usepackage{rotating}
\usepackage{tabularx}
\usepackage[breaklinks,hidelinks]{hyperref}

\hypersetup{
	colorlinks=true,
	linkcolor=blue,     % Color of internal links
	urlcolor=blue,      % Color of URLs
	citecolor=blue,     % Color of citations
	filecolor=blue      % Color of file links
}

\usepackage{breakurl} 
\usepackage{xr}
\usepackage[percent]{overpic}
\usepackage[]{caption}
\usepackage{algorithmicx}
\usepackage[]{algpseudocode}
\usepackage{algorithm}
\usepackage{diagbox}
\usepackage{graphicx}
\usepackage{wrapfig}
\usepackage{lscape}
\usepackage{fontenc}
\usepackage{setspace}
\usepackage{bm}
\usepackage{slashbox}
\usepackage{lscape}
\usepackage{breakurl} 
\usepackage{multirow,booktabs}
\usepackage{eurosym}
\usepackage{titlesec}
\usepackage{sectsty}
\usepackage{placeins}
\usepackage{subcaption}
\usepackage[thinc]{esdiff}
\usepackage{derivative}
\usepackage[flushleft]{threeparttable}
\usepackage{adjustbox}
\usepackage{mlmath}
\usepackage{algpseudocode}
\usepackage{booktabs}

\epsfverbosetrue
\def\theequation{\thesection.\arabic{equation}}  
\def\abstract{\if@twocolumn
\section*{Abstract}
\else \normalsize 
\begin{center}
{\bf Summary\vspace{-.5em}\vspace{0pt}} 
\end{center}
\quotation 
\fi}
\def\endabstract{\if@twocolumn\else\endquotation\fi}

\makeatletter
\newcommand{\myappendix}[1]{
	\setcounter{section}{1}
        \renewcommand{\thesection}{A\arabic{section}}}

\usepackage[margin=1in]{geometry}
\titlespacing*\section{0pt}{0pt plus 4pt minus 2pt}{0pt plus 2pt minus 2pt}
\titlespacing*\subsection{0pt}{0pt plus 4pt minus 2pt}{0pt plus 2pt minus 2pt}
\titlespacing*\subsubsection{0pt}{0pt plus 4pt minus 2pt}{0pt plus 2pt minus 2pt}

\definecolor{myblue}{RGB}{0,73,114}

\newcounter{mynotation}

\usepackage{paralist}

\renewenvironment{itemize}[1]{\begin{compactitem}#1}{\end{compactitem}}
\renewenvironment{enumerate}[1]{\begin{compactenum}#1}{\end{compactenum}}

\usepackage[skins,breakable]{tcolorbox}
\newtcolorbox{cvbox}[2][]{%
	blanker,
	after skip=8mm,%   enlarge distance to the next box
	title=#2,
	coltitle=cyan,
	#1
}
\makeatletter
\def\@seccntformat#1{\@ifundefined{#1@cntformat}%
	{\csname the#1\endcsname\quad}  % default
	{\csname #1@cntformat\endcsname}% enable individual control
}
\let\oldappendix\appendix %% save current definition of \appendix
\renewcommand\appendix{%
	\oldappendix
	\newcommand{\section@cntformat}{\appendixname~\thesection\quad}
}
\makeatother

\usepackage{titlesec}

\begin{document}
\setlength{\abovedisplayskip}{0.15cm}
\setlength{\belowdisplayskip}{0.15cm}
\pagestyle{empty}
 \begin{titlepage}

\title{\bfseries\sffamily\color{myblue}  
Vector Copula Variational Inference and Dependent Block Posterior Approximations}
\author{Yu Fu, Michael Stanley Smith and Anastasios Panagiotelis}
\date{\today}
\maketitle
\noindent
{\small Yu Fu is a PhD student and Michael Smith is Chair of Management (Econometrics), both at the Melbourne Business School, University of Melbourne, Australia.
Anastasios Panagiotelis is Professor of Econometrics and Business Statistics at Monash University, Australia.  Correspondence should be directed to Michael Smith at {\tt mikes70au@gmail.com}. 
\\

\noindent \textbf{Acknowledgments}: Yu Fu gratefully acknowledges support from the University of Melbourne through the Faculty of Business and Economics Graduate
	Research Scholarship and Henry Buck Scholarship. Michael Smith's research is supported 
	by Australian Research Council Discovery Project DP250101069. We thank Jordan G. Bryan, David Nott and Lin Deng for helpful comments. 
}\\

\noindent \textbf{Data Availability Statement:} The real datasets that support the findings of this study are available in the public domain at websites with URLs given in the text. All simulated datasets can be generated using the code given. 

%\normalsize
\newpage
\begin{center}
\mbox{}\vspace{2cm}\\
{\LARGE \title{\bfseries\sffamily\color{myblue} Vector Copula Variational Inference and Dependent Block Posterior Approximations}
}\\
\vspace{1cm}
{\Large Abstract}
\end{center}
\vspace{-1pt}
\onehalfspacing
\noindent
Variational inference (VI) is a popular method to estimate statistical
models.
The key to VI is the selection of a tractable density to approximate the Bayesian posterior.
For large and complex models a common choice is to assume independence between multivariate  blocks in a partition of the parameter
space. While this simplifies the problem it can reduce accuracy. This paper proposes
using vector copulas to capture dependence between the blocks parsimoniously.
Tailored multivariate marginals are constructed using learnable transport maps. We call the resulting joint distribution a ``dependent block posterior'' approximation.
Vector copula models are suggested that make tractable and flexible variational approximations. They allow for differing marginals, numbers of blocks, block sizes and forms of between block dependence. They also allow for solution of the variational optimization 
using 
efficient stochastic gradient
methods. The 
approach is demonstrated using four different statistical models and 16 datasets which have posteriors that are challenging to approximate.
This includes models that use global-local shrinkage priors for regularization, and 
hierarchical models for smoothing and heteroscedastic time series. 
In all cases, our method produces more accurate posterior approximations than benchmark VI methods that either assume block independence or factor-based dependence, at limited additional computational cost. %A python package implementing the method is available on GitHub.

%Variational inference (VI) has been a popular method to compute Bayesian posterior distributions. The key to VI is to choose a suitable variational density to capture important features from a posterior distribution. A factorized variational density is common choice when the variables in a posterior distribution can be partitioned into several blocks. In this paper, we use vector copulas~\citep{fan2023vector} to additionally capture the dependence between blocks of variables. Vector copula VI is a direct extension of factorized VI and can parameterize the dependences within and between blocks separately. We develop efficient reparameterization tricks to sample from Gaussian vector copulas and Kendall vector copulas in~\citet{fan2023vector}, respectively. The multivariate margin for each block is constructed by a fixed form cyclically monotone transformation. We show the importance of including dependence between blocks by considering two Bayesian variable selection models with a horseshoe prior and a Bayes Lasso prior, respectively. In each case, vector copula VI is scalable and increases the performance of variational approximations substantially. Additionally, two state-space models are considered to demonstrate the use of vector copulas to link multivariate margins in VI.

\vspace{20pt}
 
\noindent
{\bf Keywords}: Re-parameterization trick; Global-local shrinkage priors; Normalizing Flows; Transport maps; Variational Bayes; Vector copula
\end{titlepage}
%\doublespacing

\newpage
\pagestyle{plain}
\setcounter{equation}{0}
\renewcommand{\theequation}{\arabic{equation}}
 \section{Introduction} \label{sec1}
Variational inference (VI) methods are a scalable alternative to Markov chain Monte Carlo (MCMC) methods for computing Bayesian inference; for some recent applications in econometrics, see
\cite{koopkorobilis2023}, \cite{loaiza2023fast}, \cite{bernardi2024} and~\cite{deng2024}. VI methods minimize the divergence between the posterior density and an approximating density $q\in \fQ$.
Key to the success of VI is the selection of a distributional family $\fQ$ that 
balances approximation accuracy with the speed at which this optimization problem can be solved. For the large and complex models for which VI is favored, one popular approach is to 
partition the model parameters into blocks that are assumed independent, but with 
multivariate marginals that are tailored to the statistical model. This is called
a mean field or factorized approximation~\citep{ormerod2010explaining}. Models
where 
such approximations have been used to compute VI include deep neural networks~\citep{graves2011practical}, random effects models~\citep{goplerud2022fast, menictas2023streamlined},
vector autoregressions~\citep{bernardi2024},
state space models~\citep{quiroz2023gaussian} tobit models~\citep{danaher2020}, and topic models~\citep{ansari18} 
%and neural networks~\citep{JMLR:v20:19-236},
among others. However, this independence assumption can reduce approximation accuracy.
%However, the impact on posterior approximation accuracy of this independence assumption
%is rarely studied. 
In this paper, we use the recent idea of a vector copula~\citep{fan2023vector} 
to capture dependence between the blocks parsimoniously, while retaining tailored multivariate marginals.
We show how this can improve approximation accuracy with limited  
increase in the computation time. We refer to the approximation as a ``dependent block posterior'' and call our new method ``vector copula variational inference'' (VCVI).

Conventional copula models~\citep{nelsen06} couple (strictly) univariate marginals together to create a dependent joint distribution, and have been used previously as variational approximations~\citep{tran2015copula,han2016variational,smith2020high,gunawanetal24}. 
In particular, implicit copula models that are constructed through element-wise transformation of the model parameters---which include elliptical and skew-elliptical copulas---make highly scalable approximations~\citep{smith2020high,smith2023implicit,salomone2023structured}. They are also fast to calibrate
because they allow easy application of the re-parameterization trick~\citep{kingma2013auto} in stochastic gradient descent (SGD) algorithms, to provide 
efficient solution of the variational optimization~\citep{salimansknowles2013,titsias2014doubly}. 

Conventional copula models are based on Sklar's theorem, which cannot be employed with multivariate marginals~\citep{genest1995}. However, \cite{fan2023vector} suggest an extension 
that they call ``Vector Sklar's Theorem'', where
each block of a random vector is transformed to a multivariate uniform distribution by a 
transport map. 
A vector copula is then a distribution function on the product space of these multivariate uniform distributions. \cite{fan2023vector}~show that the density of a multivariate distribution can be expressed in terms of a vector copula for a given partition of the random vector. Motivated by this result, vector copulas can be used to combine multivariate marginals in a joint
density with inter-block dependence. 

Whereas~\cite{fan2023vector} use vector copula models in 
low dimensions to model data, we use them here in high dimensions as variational approximations. Two different learnable flexible fixed form transport maps are considered to define the
multivariate marginals of the blocks. These are coupled together by either a Gaussian vector copula (GVC) or Kendall vector copula (KVC) to specify 
dependent block variational posteriors. 
The GVC has a restricted form parameter matrix, for which 
we consider two characterizations: a factor pattern and an orthogonal pattern. 
Fast generative representations are
given for these vector copula models, which are necessary to implement SGD
with re-parameterized gradients for variational optimization.
%The modular nature of our vector copula approximations is also illustrated by considering
%vector copulas with multivariate marginals that are themselves either conventional copulas or even other vector copulas. 
 
To demonstrate the efficacy and broad applicability
of our VCVI method we employ it to estimate 
four example statistical models.
% They all have posteriors that are well approximated
%using vector copula models.
The first is a logistic regression, with coefficients regularized using a horseshoe 
prior~\citep{carvalho2010horseshoe}, for sparse signals from nine datasets with between 38 and 10,000 covariates. 
The second is a correlation matrix parameterized by spherical co-ordinates that are regularized using a Bayesian LASSO~\citep{figueiredo2003adaptive}, applied to an economic dataset with 49 variables and only 103 observations. 
The third is an unobserved component 
stochastic volatility model for U.S. inflation~\citep{chan2017stochastic}.
The fourth is an additive p-spline model 
with a hierarchical prior for regression function smoothing~\citep{lang2004bayesian}.
In all four examples,
it is attractive to partition the model parameters into blocks. We show that dependent block 
posteriors from VCVI are more accurate approximations than assuming inter-block independence. 
They also out-perform other benchmark variational approximations that capture dependence in the posterior without partitioning the parameters into blocks, including 
factor covariance Gaussians~\citep{ong2018gaussian} and conventional 
Gaussian copulas~\citep{han2016variational,smith2020high}.

A few studies have considered dependent block posteriors for VI. 
\cite{menictas2023streamlined} use different product restrictions in factorized VI to estimate a model with crossed random effects. They find that allowing global parameters to be correlated with random effects improves accuracy. Similarly,~\citet{goplerud2023partially} show that making fixed effects dependent on random effects in a generalized linear mixed model can improve estimation of the posterior in VI. \citet{loaiza2023fast} consider a Gaussian approximation with a parsimonious patterned covariance matrix for inter- and intra-block dependence exhibited by the posterior of a multinomial probit model. 
However, as far as we are aware, our study is the first to construct dependent block posterior 
approximations in VI in a general fashion using vector copulas.

The rest of the paper is organized as follows. Section~\ref{sec2} briefly introduces VI using SGD  and copula models, and then outlines vector copula models and their use as variational approximations.
 Section~\ref{sec3} suggests vector copulas that are both attractive choices to capture dependence in blocked posteriors and also learnable using SGD with the re-parameterization trick.
 Section~\ref{sec4} contains the four example applications of VCVI, while Section~\ref{sec5} concludes. An
 Appendix discusses the Kendall's vector copula, and a five part Online Appendix provides derivations, implementation details and further empirical results. A Python package that implements our method, and code to replicate the results in this paper, can be found at \href{https://github.com/YuFuOliver/VCVI_Rep_PyPackage}{{\tt https://github.com/YuFuOliver/VCVI\_Rep\_PyPackage}}.

 % !TeX spellcheck = en_US
\setlength{\abovedisplayskip}{1pt}
\setlength{\belowdisplayskip}{1pt}
\section{Vector Copula Variational Inference}\label{sec2}
In this section we first review the basic concepts of VI and copula models, and then
introduce vector copula models and their use as a variational approximation (VA). Suitable choices of learnable vector copulas are discussed later in Section~\ref{sec3}.

\subsection{Variational Inference}
Let $\vtheta\in \sR^d$ be a vector of unknown parameter values, possibly augmented with some latent variables, and $\vy$ be
 observed data. 
VI approximates the posterior $p(\vtheta | \vy) \propto p(\vy | \vtheta) p(\vtheta) \equiv h(\vtheta)$, where $p(\vy | \vtheta)$ is the likelihood and $p(\vtheta)$ is the prior density, by a VA with 
density $q(\vtheta)\in {\cal Q}$. To learn the VA it is common to minimize the Kullback-Leibler 
divergence (KLD) between $q(\vtheta)$ and $p(\vtheta | \vy)$. This is equivalent to maximizing the evidence lower bound (ELBO) \citep{ormerod2010explaining}
\begin{equation}\label{eq: elboq}
	\fL(q) = \int \log \frac{h(\vtheta)}{q(\vtheta)} q(\vtheta) d \vtheta = \Exp_{q}\left(\log h(\vtheta)-\log q(\vtheta)\right)\,,
\end{equation}
over $q\in {\cal Q}$ which is called the variational optimization. 

Key to effective VI is
the selection of a
distributional family ${\cal Q}$
that provides an accurate approximation, while also allowing fast solution of 
the variational optimization. When $q(\vtheta)$ is parameterized by $\vlambda$
(a so-called ``fixed form'' VA) we write the density as $q_\lambda(\vtheta)$
and solve the optimization problem by SGD
combined with the ``re-parameterization trick"~\citep{kingma2013auto,rezende2014}. 
This trick draws a random vector $\vepsilon\sim f_\epsilon$, such that $\vtheta = g(\vepsilon, \vlambda)\sim q_\lambda$
for a deterministic function $g$, so that~\eqref{eq: elboq} can be written as
\begin{equation}\label{eq: elbolambda}
	\fL(\vlambda) = \Exp_{f_\epsilon}\left( \log h \left( g\left( \vepsilon, \vlambda \right)\right)-\log q_\lambda \left( g\left( \vepsilon, \vlambda \right)\right) \right)\,,
\end{equation}
%
%reduce the noise in the gradient. It samples a random vector $\vepsilon$ from a base distribution $f(\vepsilon)$, and then assigns a deterministic path to generate $\vtheta$ such that $\vtheta = g(\vepsilon, \vlambda)$, where $\vlambda$ are the variational parameters controlling the variational density $q(\vtheta)$. As a result, the ELBO function from (\ref{eq: elboq}) can be expressed as:
%\begin{equation}\label{eq: elbolambda}
%	\fL(\vlambda) = \Exp_{f}\left( \log h \left( g\left( \vepsilon, \vlambda \right)\right)-\log q_{\vlambda} \left( g\left( \vepsilon, \vlambda \right)\right) \right).
%\end{equation}
where we now write the ELBO as a function of $\vlambda$.
Differentiating under the integral in~\eqref{eq: elbolambda} gives:
\begin{equation}\label{eq: elbogradient}
	\nabla_{\vlambda} \fL (\vlambda) = \Exp_{f_\epsilon} \left\{ \left[ \frac{dg(\vepsilon, \vlambda)}{d \vlambda} \right]^\top \nabla_{\vtheta} \left( \log h(\vtheta) - \log q_\lambda(\vtheta)\right) \right\} .
\end{equation}
For careful choices of $q_\lambda$, $f_\epsilon$ and $g$, a Monte Carlo estimate of the expectation in~\eqref{eq: elbogradient} has low variability.
%~\citep{xu2019variance}.
Usually, replacing the expectation with a single draw $\vepsilon \sim f_\epsilon$ is sufficient (as we do here), making optimization using SGD
fast.\footnote{It is usual to refer to this method as a descent algorithm, even though the variational optimization is a maximization problem.} 
The elements of $\vlambda$ are either unconstrained or transformed to be so\footnote{While we do not explicitly mention these transformations in the rest of the text, we replace $\lambda_i$ with $\tilde \lambda_i\in \sR$ in the following circumstances: (i)~if $\lambda_i>0$ we set $\lambda_i = \tilde{\lambda}_i^2$, and (ii)~
if $\lambda_i\in (a,b)$ we set $\lambda_i = a + (b-a)/(1 + \exp(-\tilde{\lambda}_i))$.} and it is unnecessary for $\vlambda$ to be uniquely identified.
 
For complex and/or large statistical models it is popular to partition
$\vtheta=(\vtheta_1^\top,\vtheta_2^\top,\ldots,\vtheta_M^\top)^\top$ and assume independence between
blocks, so that $q(\vtheta)=\prod_{j=1}^M q_j(\vtheta_j)$. This is called a factorized or mean field
approximation, and it simplifies the solution of the optimization (e.g. using coordinate ascent algorithms as in~\cite{ormerod2010explaining}) while allowing selection
of approximations
$q_j$ to match the marginal posteriors $p(\vtheta_j|\vy)$. However, this independence assumption can increase approximation error substantially for some posteriors. 
In this paper we use a vector copula model for $q$ to allow for tailored marginals $q_j$,
while also introducing posterior dependence between the blocks. We show how to use SGD with re-parameterization gradients at~\eqref{eq: elbogradient} to provide fast solution of the optimization. 
 
% will be used as the stochastic gradient and passed to a stochastic optimization algorithm, such as Adadelta~\citep{zeiler2012adadelta} and Adam~\citep{kingma2014adam}. 

% -------- consider the following in the introduction part --------------------------------------
%The use of the reparameterization trick, as in (\ref{eq: elbolambda}) and (\ref{eq: elbogradient}), has greatly facilitated the research of VI with a general dependence structure in the variational density. For example,~\citet{titsias2014doubly} outline a general method to optimize a lower-triangular matrix, which depicts the covariance matrix of the variational density. Followed by their work,~\citet{tan2018gaussian} use a sparse precision matrix to represent the conditional independence structure in state-space models and longitudinal random effect models. Additionally,~\citet{ong2018gaussian} specify $q(\vtheta)$ as a Gaussian distribution with a factor covariance matrix, which has been the base model for several extensions~\citep{smith2020high, loaiza2022fast}. On the other side, a factorized variational density is still of interest......

%~\citet{tran2015copula} extend the MFVI by using a copula to link marginal distributions $q(\theta_i)$, such that $q(\vtheta) = c \left( F_1(\theta_1), \dots, F_M(\theta_M)\right) \prod_{i = 1}^{M} q(\theta_i)$, where $c(\cdot)$ is a multivariate copula density and $F_i(\theta_i)$ is the cumulative distribution function (cdf) of $\theta_i$. 

% ---------------------------------------------------------------------------------------------
\subsection{Conventional copula models} \label{sec: copula model}
Before introducing vector copulas, we briefly outline conventional copula models and their use as VAs.
These are motivated by Sklar's theorem,
which represents a distribution by its univariate marginals and a copula
function \citep[pp.42-43]{nelsen06}. For continuous
$\vtheta = (\theta_1, \dots, \theta_d)^\top$, a copula model has density
\begin{equation}\label{eq: copuladensity}
	q(\vtheta) = c\left(F_1(\theta_1), \dots, F_d(\theta_d)\right) \prod_{i = 1}^{d} q_i(\theta_i)\,,
\end{equation}
where $F_i$ and $q_i=\frac{\partial F_i}{\partial \theta_i}$ are marginal distribution and density functions of $\theta_i$, respectively, 
and $c(\vu)$ is a density function on $\vu=(u_1,\ldots,u_d)^\top\in [0,1]^d$ with uniform marginals, called the copula density. A copula model is a flexible way of constructing a multivariate distribution through choice of $c$ and $q_1,\ldots,q_d$;
%Starting with a random vector $\mathbf{u} = (u_1, \ldots, u_d)^\top \in[0,1]^d$ with distribution function given by the copula, each element $u_i$ is transformed to $\theta_i$ by a monotonic transformation $t_i$ such that $\theta_i = t_i(u_i)$. Equation~(\ref{eq: copuladensity}) can be rewritten by changing variables so that $q(\vtheta) = c(u_1, \dots, u_d) \prod_{i = 1}^{d} \frac{\partial t_i^{-1}(\theta_i)}{\partial \theta_i}$.
%The variables thus transformed will have a marginal distribution functions given by the inverse of the corresponding monotonic transformation. 
see~\cite{nelsen06} and~\cite{joe2014dependence} for introductions.

\cite{tran2015copula,han2016variational,smith2020high,smith2023implicit} and \cite{gunawanetal24} use copula models as VAs. However, evaluation 
of the gradient $\nabla_\theta \log q(\vtheta)$
in~\eqref{eq: elbogradient} can be too slow for some choices of copula $c$ and/or marginals $q_i$. 
To solve this problem, \cite{smith2023implicit} construct copula models by element-wise monotonic transformations $\theta_i=t_i(u_i)$, so that 
$q_i(\theta_i)=\frac{\partial t_i^{-1}(\theta_i)}{\partial \theta_i}$ and $u_i=F_i(\theta_i)=t_i^{-1}(\theta_i)$ at~\eqref{eq: copuladensity}.
%To solve this problem, \cite{smith2020high} show that copula models formed by element-wise monotonic transformations of $\vtheta$ can provide an expression for $\log q$ with a gradient that is fast to compute.
%For example,~\cite{smith2023implicit} consider
%fixed form monotone bijective transformations from an elliptical distribution to construct an elliptical copula model. It has an alternative expression for $q(\vtheta)$ that allows easy application of the re-parameterization trick and 
%fast solution of the optimization by SGD.
For example, for an elliptical copula~\citep{frahm2023} these 
authors use the learnable transformations
\begin{equation}\label{eq:unitrans}
	\theta_i=t_i(u_i)\equiv b_i + s_i \cdot k_{\veta_i}\left( F_e^{-1}(u_i) \right)\,,\;\,\mbox{ for }i=1,2,\ldots,d\,.
\end{equation} 
Here, $b_i$ and $s_i$ capture the mean and scale of $\theta_i$, respectively, $F_e^{-1}$ is the quantile function of a univariate standardized elliptical distribution (e.g. $\Phi^{-1}$ for the Gaussian), and $k_{\veta_i}$ is a monotone function with parameters $\veta_i$ that captures the marginal asymmetry of $\theta_i$. For careful choices 
%the variational
%parameters are $\vlambda=\{(b_i,s_i,\veta_i);i=1\ldots,d\}\cup \vlambda_c$, and for careful choices
of $k_{\veta_i}$, evaluation of the required gradient is fast.
%; see~\cite{smith2023implicit}
%for a detailed discussion. 
We show in Section~\ref{sec:VDVC}
that a generalization of this transformation approach can also be used to specify vector copula models as VAs.

\subsection{Vector copula models}\label{sec:vcm}
Sklar's theorem is strictly defined for univariate marginals, and it is 
impossible to simply replace the arguments of the copula density in~\eqref{eq: copuladensity} with 
multivariate marginals $F_j$; see~\cite{genest1995,ressel2019}. However,~\citet{fan2023vector} developed
the idea of a vector copula that is based on an extension of Sklar's theorem, as we now outline.

Consider the partition $\vtheta=(\vtheta_1^\top,\ldots,\vtheta_M^\top)^\top$, where
$\vtheta_j\in \sR^{d_j}$ and $d = d_1 + \dots + d_M$. An $M$-block vector copula $C_v$ is defined as a distribution function on $[0,1]^d$ with multivariate uniform marginal distributions $\mu_j$ on $[0,1]^{d_j}$,
for $j=1,\ldots,M$.\footnote{\cite{fan2023vector} do not label $C_v$ as ``$M$-block'', but we do so here as we find it lends clarity.} 
There is no ``within dependence'' in $\mu_j$, but $C_v$ captures the ``between dependence'' of the $M$ blocks. 
Theorem~1 of~\cite{fan2023vector} gives the vector extension of Sklar's theorem, and we focus on its use for an absolutely continuous probability distribution $P$ on $\sR^{d_1}\times \cdots \times \sR^{d_M}$, because this is the case for our VA. If $P_j$ on $\mathbb{R}^{d_j}$ is the $j$th multivariate marginal of $P$, then there is a transformation $\nabla \psi_j: [0,1]^{d_j} \rightarrow \sR^{d_j}$, where $\nabla \psi_j$ is the gradient of a convex function $\psi_j$, that pushes forward $\mu_j$ to $P_j$ (i.e. $(\nabla \psi_j)_{\#}\mu_j = P_j$). Moreover, the density function of $P$ is
\begin{equation}\label{eq: vcdensity}
%	\begin{aligned}
		q(\vtheta) = c_{v}\left(\nabla \psi_1^*(\vtheta_1), \dots, \nabla \psi_M^*(\vtheta_M)\right)\ \prod_{j=1}^{M}q_j(\vtheta_j) \\
		= c_{v}(\vu_1, \dots, \vu_M) \prod_{j=1}^{M}q_j(\vtheta_j),
%	\end{aligned}
\end{equation}
\noindent where $\psi_j^*$ is the convex conjugate of $\psi_j$, 
$\vu_j =\nabla \psi_j^*(\vtheta_j)\in [0,1]^{d_j}$, $\vu=(\vu_1^\top,\ldots,\vu_M^\top)^\top\sim C_v$,
and $c_{v}(\vu)=\frac{\partial^d}{\partial u_1\cdots\partial u_d}C_v(\vu)$ is 
an $M$-block vector copula density. For absolutely continuous $P$, $\nabla \psi_j^*=(\nabla \psi_j)^{-1}$ is the pullback of $P_j$ to $\mu_j$. %which is the case when we use~\eqref{eq: vcdensity} as a VA. %Additionally,
%if $P$ has finite second moment, then $\nabla \psi_j$ is a cyclically monotone function, although~\eqref{eq: vcdensity} still defines a valid density whenever $\nabla \psi_j$ is a 
%measure-preserving Knothe–Rosenblatt (i.e. triangular) transport map. 
When $p_j=1$ for all $j$, then $\nabla \psi_j^*=F_j$ and the density at~\eqref{eq: vcdensity} equals that at~\eqref{eq: copuladensity},
and the vector copula model nests the conventional copula model.

Equation~\eqref{eq: vcdensity} motivates the construction of learnable vector copula models
as fixed form VAs. These use the multivariate marginals 
discussed below, and the vector copulas that are discussed separately in Section~\ref{sec3}.

%An important observation is that the gradient of a convex function is a
%cyclically monotone function, and vice versa.
%A cyclically monotone function generalizes the idea of monotonicity to multivariate functions and is defined as follows. 
%A function $T: \sR^{d_j} \to \mathbb{R}^{d_j}$ is cyclically monotone if for any finite sequence of points $\vx_1, \vx_2, \ldots, \vx_k$ in its domain, 
%$\sum_{i=1}^{k} T(\vx_i)^\top (\vx_{i+1} - \vx_i) \leq 0 $
%and $\vx_{k+1} = \vx_1$ to complete the cycle. Moreover, compositions of cyclically monotone functions are also cyclically monotone. This motivates the construction 
%of multivariate marginals from such functions as we discuss below. 

%
%\cite{fan2023vector} also note that it is generally hard to construct parametric vector copulas with a single measure push-forward, and instead
%employ a composite measure transport as follows. For $n_j\geq 1$, let $T_j^* = \nabla \psi_{j,1}^* \circ \nabla \psi_{j,2}^* \circ \dots \circ \nabla \psi_{j,n_j}^*$ be a composite function such that $T_j^*:\sR^{d_j} \rightarrow [0,1]^{d_j}$, then simply replace $\nabla \psi_j^*$ by
%$\vu_j=T_j^*(\vtheta_j)$ in
%\eqref{eq: vcdensity}. Because the composite measure transport of cyclically monotone functions is also cyclically monotone, multivariate marginals can be readily constructed 
%from such functions as we now discuss.

\subsection{Multivariate marginal approximations}\label{sec:VDVC}
In this paper, we replace $\nabla \psi_{j}^*$ in~\eqref{eq: vcdensity} with the inverse of a
learnable fixed form continuously differentiable bijective transformation $T_j:[0,1]^{d_j}\rightarrow \mathds{R}^{d_j}$, for each $j=1,\ldots,M$. If $T_j$ is cyclically monotone, then this matches the vector copula formulation above. Nevertheless, when $T_j$ is not cyclically monotone,~\eqref{eq: vcdensity} still gives a well-defined 
density with the same within and between block dependence decomposition.  
In the 
vector copula model, 
$\vu_j\sim \mu_j$ is uniformly distributed and 
$\vtheta_j = T_j(\vu_j)$, so that
$T_j$ solely determines the multivariate marginal of $\vtheta_j$. 
By considering a change of variables from 
 $\vtheta_j$ to $\vu_j$, the marginal density is given by
\[q_j(\vtheta_j)=\left| \mbox{det} \left\{ \frac{\partial T_j^{-1}}{\partial \vtheta_j} \right\} \right|\,.\] 
Two bijective transformations $T_j$ that provide a great deal of flexibility in constructing $q_j$ are discussed below. 

The first transformation is a generalization of that at~\eqref{eq:unitrans} 
to the multivariate case, and is similar to that used by~\citet{salomone2023structured}: 
\begin{equation}\label{eq: mmm1}
\mbox{M1}:\;\;\;\;\;\;\;	T_j(\vu_j) = \vb_j + S_j \underline{k}_{\veta_j}\left\{ L_j \underline{\Phi}^{-1}(\vu_j) \right\}. 
\end{equation}
Here, $\vb_j$ captures location, $S_j$ is a diagonal matrix with positive elements to capture scale, and $L_j$ is a lower triangular matrix with ones on its diagonal to capture within dependence. 
The function $\underline{\Phi}^{-1}(\vu_j)$ is the standard Gaussian quantile function applied element-wise to $\vu_j$. If 
$\vx_j=(x_1,\ldots,x_{d_j})^\top$ and $\veta_j=(\veta_{j,1}^\top,\ldots,\veta_{j,d_j}^\top)^\top$, then the function 
$\underline{k}_{\veta_j}(\vx_j)=(k_{\veta_{j,1}}(x_1),\ldots,k_{\veta_{j,d_j}}(x_{d_j}))^\top$ with $k_{\veta_{j,s}}:\sR \rightarrow \sR$ a monotonic function parameterized
by $\veta_{j,s}$. The variational parameters for marginal $q_j$ are $\vlambda_j = (\vb_j^\top,\mbox{diag}(S_j)^\top,\mbox{vech}(L_j)^\top,\veta_j^\top)^\top$, where 
$\mbox{vech}(L_j)$ denotes the half-vectorization of $L_j$ for only the non-fixed lower triangular elements. We refer to the transformation at~\eqref{eq: mmm1} and its resulting marginal distribution as ``M1''.

For $k_{\veta}$ we use the inverse of the YJ transformation~\citep{yeo2000new}, which is parameterized by a scalar $0 < \eta < 2$, such that
\begin{equation}\label{eq: IYJ}
	k_{\eta}(x) =
	\begin{cases}
			1 - \left( 1 - x(2 - \eta) \right)^{\frac{1}{2 - \eta}} & \text{if } x < 0, \\
			\left( 1 + x\eta \right)^{\frac{1}{\eta}} - 1 & \text{if } x \geq 0\,,
		\end{cases}
\end{equation}
\noindent and equals the identity transformation $k_{\eta}(x) = x$ when $\eta = 1$. \cite{smith2020high} show this transformation
is effective in capturing skewed marginals in $\theta_i$, 
although other choices for $k_{\veta}$ include the G\&H transformation of Tukey %~\citep{headrick2008} 
and the  
inverse of the sinh-arcsinh transformation of~\cite{jones2009} used by~\cite{salomone2023structured}. 

%The parameters of the transformation at~\eqref{eq: mmm1} are $\vlambda_j \equiv \{\vb_j,\mbox{diag}(S_j),\mbox{vech}(L_j),\veta_j\}$, and they parameterize the 
%marginal $q_j$ of the VA. 

As we demonstrate later, an attractive feature of M1 is that a pattern may be adopted for $L_j$ that is a suitable for $p(\vtheta_j|\vy)$,
reducing the 
number of variational parameters $\vlambda_j$ and improving performance of the SGD algorithm when $d_j$ is high. 
For example, in our empirical work we constrain $L_j$ to be diagonal when within dependence of $\vtheta_j$ is close to zero, and $L_j^{-1}$ to be a band matrix when an ordering of the elements of $\vtheta_j$ is serially dependent.

When $d_j$ is large and a patterned $L_j$ is inappropriate, then we adopt the transformation
\begin{equation}\label{eq:mmm2}
\mbox{M2}:\;\;\;\;\;\;\;	T_j(\vu_j) = \vb_j + \underline{k}_{\veta_j}\left(  E_j \underline{\Phi}^{-1}(\vu_j) \right). 
\end{equation}
Here, $E_j = J_jJ_j^T + D_j^2$, where $J_j$ is a $(d_j \times w)$ matrix with $w < d_j$,  $D_j$ is a diagonal matrix, and $\vb_j$, $\underline{\Phi}^{-1}(\vu_j)$, $\underline{k}_{\veta_j}$ are as defined as before.  When $d_j$ is large and $w<<d_j$ this transformation 
captures within dependence more parsimoniously than does M1 with dense $L_j$.  The variational parameters are $\vlambda_j = (\vb_j^\top, \mbox{vec}(J_j), \mbox{diag}(D_j), \veta_j^\top)^\top$, and the dimension of $\vlambda_j$ increases only linearly with $d_j$.
The determinant and inverse of $E_j$ can be computed efficiently by the Woodbury formula, which is necessary when evaluating the gradient of $q$. We refer to this transformation and its resulting marginal distribution as ``M2''.

We make five observations on these multivariate marginals. First, when the identity transformation $\underline{k}_{\veta_j}(\vx_j)=\vx_j$ is employed (i.e. where $\eta=1$ in~\eqref{eq: IYJ} for each element) then M1 and M2 correspond to multivariate Gaussians $\vtheta_j\sim N(\vb_j,S_jL_jL_j^\top S_j)$ and $\vtheta_j\sim N(\vb_j,E_jE_j^\top)$, respectively. Second, when 
$\veta_{j}$ is learned from the data, then M1 and M2 correspond to Gaussian copula models with flexible skewed marginals as shown in~\cite{smith2020high}.
Third, if $\underline{\Phi}^{-1}$ is replaced in~\eqref{eq: mmm1} and~\eqref{eq:mmm2}
with the element-wise application of the quantile function of another 
elliptical distribution, then M1 and
M2 correspond to elliptical copula models~\citep{smith2023implicit}.
Fourth, other learnable bijective transformations can also be used here, such as the 
cyclically monotone maps discussed by~\cite{bryan2021multirank} and~\cite{makkuva2020optimal}. The fifth observation is that M1 and M2 are types of normalizing 
flows~\citep{papamakarios2021normalizing} with Gaussian base distribution  $\vz_j=\underline{\Phi}^{-1}(\vu_j)\sim N(\bm{0},I_{d_j})$. Normalizing flows are employed
widely as effective distributional approximations, as we find for M1 and M2.

%AP: I REMOVED THE PARAGRAPH BELOW BECAUSE IT IS ALL REPETITION.

%Finally we note that when $d_j=1$, $T_j^{-1}$ is the monotone distribution function $F_j$, and the density at~\eqref{eq: copuladensity} equals that at~\eqref{eq: vcdensity},
%so that the vector copula model nests the conventional copula model. Moreover, 
%constructing the multivariate marginal by selecting a learnable cyclically monotone function $T_j$ 
%generalizes the construction of a univariate marginal of $\theta_i$ by selecting a learnable monotone function $t_i$ as in Section~\ref{sec: copula model}.

%AP: IS THE ANY REASON WHY THIS SUBSECTION WOULD NOT BE BETTER PLACED AT THE END OF 3?

\subsection{Variational optimization}
We evaluate~\eqref{eq: elbolambda} using the re-parameterization trick
with $g$ for the vector copula model given by the composition of two functions. The first is
$\vu=g_1(\vepsilon, \vlambda_{\mbox{\footnotesize vc}})\sim C_v$, where $\vepsilon\sim f_\epsilon$ and $\vlambda_{\mbox{\footnotesize vc}}$ 
are the vector copula parameters, which is specified separately for different vector copulas below in Section~\ref{sec3}. The second function is $g_2(\vu, \vlambda_{\mbox{\footnotesize marg}}) = (T_1(\vu_1; \vlambda_1)^\top, \dots, T_M(\vu_M; \vlambda_M)^\top)^\top$, where $\vlambda_{\mbox{\footnotesize marg}} = (\vlambda_1^\top, \cdots, \vlambda_M^\top)^\top$
denotes the parameters of all $M$ marginals and are simply the parameters of the transformations M1 or M2 discussed above. 
Then setting  $\vlambda = ( \vlambda_{\mbox{\footnotesize vc}}^\top,  \vlambda_{\mbox{\footnotesize marg}}^\top)^\top$, the transformation is
\begin{equation*}%\label{eq: vcrep}
	\vtheta=g(\vepsilon,\vlambda):= g_2\left(g_1\left(\vepsilon, \vlambda_{\mbox{\footnotesize vc}}\right), \vlambda_{\mbox{\footnotesize marg}}\right)\,.
\end{equation*}

%For the vector copula model the re-parameterization gradient at~\eqref{eq: elbogradient}
%simplifies to
%\begin{eqnarray*} 
%	\nabla_{\vlambda_{\mbox{\footnotesize vc}}} \fL (\vlambda) &=& \Exp_{f_\epsilon}\left\{\left[ \frac{dg_2(\vu, \vlambda_{\mbox{\footnotesize marg}})}{d \vlambda_{\mbox{\footnotesize vc}}} \right]^\top \nabla_{\vtheta} \left( \log h(\vtheta) - \sum_{j=1}^{M}\log q_j(\vtheta_j) \right)  - \left[ \frac{dg_1(\vepsilon, \vlambda_{\mbox{\footnotesize vc}})}{d \vlambda_{\mbox{\footnotesize vc}}} \right]^\top \nabla_{\vu} \left( \log c_v(\vu) \right)\right\}\\
%	\nabla_{\vlambda_j} \fL (\vlambda) &= &\Exp_{f_\epsilon} \left\{\left[ \frac{dT_j(\vu_j; \vlambda_j)}{d \vlambda_j} \right]^\top \nabla_{\vtheta_j} \left( \log h(\vtheta) - \log q_j(\vtheta_j) \right)\right\}\,.
%\end{eqnarray*}

\begin{algorithm} [H]
	\caption{: {\em Vector Copula Variational Inference (VCVI) with Re-parameterized Gradient}}
	\label{alg: VCVI}
	\begin{algorithmic}
		\State Set $s=0$ and initialize $\vlambda^{(0)}$ to a feasible value
		\Repeat
		\State 1. Generate $\vepsilon^{(s)}\sim f_\epsilon$ and set $\vu = g_1(\vepsilon, \vlambda_{\mbox{\footnotesize vc}}^{(s)})$.
		\State 2. Set $\vtheta_j = T_j(\vu_j; \vlambda_j^{(s)})$ for $j = 1, \dots, M$.
		\State 3. Evaluate gradient estimate $\widehat{\nabla_\lambda \fL(\vlambda)}$ at $\vlambda=\vlambda^{(s)}$.
		\State 4. Update $\vlambda^{(s+1)}=\vlambda^{(s)}+\vrho_s \circ \widehat{\nabla_\lambda \fL(\vlambda^{(s)})}$ where $\vrho_s$ is a vector of adaptive step sizes.
		\State 5. Set $s = s + 1$.
		\Until{Either a stopping rule is satisfied or a fixed number of steps is taken.}
	\end{algorithmic}
\end{algorithm}
Algorithm~\ref{alg: VCVI} applies SGD with the re-parameterization trick using the generative representation $g$ specified above to solve the variational optimization. Step~1 depends on the choice of vector copula, while Step~2 on the choice of multivariate marginal. At Step~3 an unbiased estimate of the re-parameterized gradient at~\eqref{eq: elbogradient} is obtained by replacing $\vepsilon$ (therefore also $\vtheta=g(\vepsilon,\vlambda)$) by its single drawn value $\vepsilon^{(s)}$. In Step~4, ``$\circ$'' denotes the Hadamard 
(i.e. element-wise) 
product, and $\vrho_s$ is the vector of  step sizes determined automatically using 
an adaptive algorithm, such as ADAM~\citep{kingma2014adam} or ADADELTA~\citep{zeiler2012adadelta}.

\section{Learnable Vector Copulas}\label{sec3}
\citet{fan2023vector} consider vector copulas for modeling data, and this section outlines two that are suitable for VCVI: the Gaussian  (GVC) and Kendall (KVC) vector copulas. Consistent with
Section~\ref{sec:vcm}, these are defined for $M$ blocks of dimension $d = \sum_{j = 1}^{M} d_j$ on $\vu=(\vu_1^\top,\ldots,\vu_M^\top)^\top$ with $\vu_j=(u_{j,1},u_{j,2},\ldots,u_{j,d_j})^\top\in [0,1]^{d_j}$. We propose
parameterizations of these vector copulas that can provide accurate approximations and allow the variational optimization to be solved efficiently. 

\subsection{Gaussian vector copula}
The GVC has distribution function
\begin{equation}\label{eq: gvccdf}
		C_v^{Ga}(\vu_1, \ldots ,\vu_M; \Omega) \equiv \Phi_d \left(  \underline{\Phi}^{-1} (\vu_1), \ldots   ,\underline{\Phi}^{-1} (\vu_M); \Omega \right)\,.
\end{equation}
Here, $\underline{\Phi}^{-1}$ denotes element-wise application of the standard normal quantile function, and $\Phi_d(\cdot;\Omega)$ is the distribution function of a $d$-dimensional zero mean normal with correlation $\Omega$. Unlike a conventional
Gaussian copula, $C_v^{Ga}$
captures no within dependence in 
each block. The correlation matrix $\Omega$ has identity matrices $I_{d_1}, I_{d_2}, \dots, I_{d_M}$ on its leading diagonal, and we denote the set of such patterned correlation matrices as ${\cal P}$. Differentiating (\ref{eq: gvccdf}) with respect to $\vu$ gives the density
\begin{equation}\label{eq: gvcpdf}
	c_v^{Ga}(\vu_1, \dots, \vu_M; \Omega) 
	= \det(\Omega)^{-1/2} \exp\left\{ -\frac{1}{2} \underline{\Phi}^{-1}(\vu)^\top \left( \Omega^{-1} - I_d \right) \underline{\Phi}^{-1}(\vu) \right\}\,.
\end{equation}
As with the conventional Gaussian copula, to sample $\vu$ from~\eqref{eq: gvccdf}, draw 
$\vz\sim N_d(\vzero,\Omega)$, then set $\vu = \underline{\Phi}(\vz)$ which denotes
the standard Gaussian distribution function applied element-wise to $\vz$. 
 
The main challenge in employing the GVC as a VA is to specify an appropriate parameterization of $\Omega \in \fP$. %This requires $\Omega$ to be positive definite and simultaneously have the required pattern. 
To be an effective VA when $d$ is large, the parameterization needs to (i)~be parsimonious, (ii)~allow
application of the re-parameterization trick for fast optimization using SGD, and (iii)~enable efficient evaluation of $\Omega^{-1}$
and $\mbox{det}(\Omega)$ when 
computing~\eqref{eq: gvcpdf} and its derivatives.
Two such parameterizations of $\Omega$ are now given.

\subsubsection{Factor pattern correlation matrix}\label{sec:GVCFp}
The first parameterization was suggested by~\citet{ren2019adaptive} in a different context.
For a scalar $\zeta > 0$ and a $(d \times p)$ matrix $B$ with $p < d$ , let
\begin{equation*}\label{eq: Omega1}
		\widetilde{\Omega} = \zeta I_d  + BB^\top\,, \quad \text{and} \quad
\Omega = A \widetilde{\Omega} A^\top\,.
\end{equation*}
Here, $A=\left[\mbox{bdiag}(\widetilde{\Omega})\right]^{-1/2}$ is the inverse of the lower triangular Cholesky factor of the matrix $\mbox{bdiag}(\widetilde{\Omega})$, which in turn denotes a block diagonal matrix with blocks of sizes $d_1, \ldots, d_M$
comprising the same elements as $\widetilde{\Omega}$ (i.e. this operator simply returns $\widetilde{\Omega}$, but with all blocks on the main diagonal to be identity matrices).
%AP: I AM NOT SURE IF THIS IS CORRECT. IS IT NOT THE CASE THAT OFF DIAGONAL BLOCKS ARE ALSO ALTERED WHEN PREMULTIPLIED BY A?
It is easy to show that  $\Omega \in \fP$. This is a factor patterned correlation matrix that captures between dependence based on a low-rank representation when $p << d$.

It is straightforward to generate $\vz\sim N_d(\vzero,\Omega)$ 
using the decomposition $\Omega = A\left( \zeta I_d  + BB^\top\right)A^\top = \zeta A A^\top + ABB^TA^\top$ as follows. 
Draw independent normals $\vepsilon = (\vepsilon_1^\top, \vepsilon_2^\top)^\top \sim N_{d+p}(\vzero, I_{d+p})$, and set
\begin{equation}\label{Omega1_rep}
	\vz  = \zeta^{1/2}A \vepsilon_1 + AB \vepsilon_2= \underline{\Phi}^{-1}(g_1(\vepsilon, \vlambda_{\mbox{\footnotesize vc}} ))\,.
\end{equation}
with $\vlambda_{\mbox{\footnotesize vc}} = (\zeta, \mbox{vec}(B)^\top)^\top$. Equation~\eqref{Omega1_rep} also provides the specification of $g_1$ for implementation of the re-parameterization trick. 
The inverse $\Omega^{-1} = A ^{-T} \widetilde{\Omega}^{-1} A^{-1}$, where  $\widetilde{\Omega}^{-1}$ can be computed efficiently
using the Woodbury formula, and $A^{-1}$ can be evaluated
using $p$ sequential rank-1 Cholesky factor updates.\footnote{By the Woodbury formula, $\widetilde{\Omega}^{-1}=\frac{1}{\zeta}[I_d-B(\frac{1}{\zeta}I_p+B^\top B)^{-1}B^\top]$ requiring inversion of only a $p\times p$ matrix. To compute the Cholesky factor $A^{-1}$ note that $\zeta I_d + BB^\top = \zeta I_d + \vb_1\vb_1^\top + \ldots +\vb_p\vb_p^\top$ where $\vb_i$ is the $i$th column of $B$, so its Cholesky factor can be computed using $p$ sequential rank-1 updates, each of $O(d^2)$.}
We label the GVC with this factor patterned $\Omega$ with $p$ factors as ``GVC-F$p$''. 

\subsubsection{Orthogonal pattern correlation matrix}
The second parameterization of $\Omega$ was suggested by~\citet{bryan2021multirank} in a different context for $M=2$ blocks, where 
\begin{equation}\label{eq: Omega2}
	\Omega = \begin{bmatrix}
		I_{d_1} & Q_1 \Lambda Q_2^\top \\
		Q_2 \Lambda Q_1^\top & I_{d_2}
	\end{bmatrix}\,.
\end{equation}
Here, $Q_1, Q_2$ are left semi-orthogonal matrices with dimensions $(d_1 \times \tilde{d})$ and $(d_2 \times \tilde{d})$, respectively, $\tilde{d} = \min{(d_1,d_2)}$, and $\Lambda=\diag(l_1,\ldots,l_{\tilde{d}})$ is a $(\tilde{d} \times \tilde{d})$ diagonal matrix. Thus, $Q_1^\top Q_1 = Q_2^\top Q_2 = I_{\tilde{d}}$. Without loss of generality,
assume $d_1 \geq d_2$, so that $\tilde{d} = d_2$ and $Q_2$ is a square orthonormal matrix.
Because $\Omega$ is a correlation matrix, then $0<\det(\Omega)<1$. The determinant of~\eqref{eq: Omega2} can be used to identify constraints
on $\Lambda$ that ensures $\Omega\in {\cal P}$ as follows. 
Let $\Omega/I_{d_1} = I_{d_2} - Q_2 \Lambda Q_1^\top I_{d_1}^{-1} Q_1 \Lambda Q_2^\top = I_{d_2} - Q_2 \Lambda^2 Q_2^\top$ be the Schur complement of block $I_{d_1}$ in the matrix $\Omega$. Then,
\begin{equation*}\label{eq: Omega2det}
	\begin{aligned}
		\det(\Omega) = &\det\left(I_{d_1}\right) \cdot \det\left(\Omega/I_{d_1}\right) 
		=   \det\left(I_{d_2} - Q_2 \Lambda^2 Q_2^\top\right) ~\\
		=& \det\left(Q_2 \left(I_{\tilde{d}} - \Lambda^2\right) Q_2^\top  \right)
		= \det(I_d - \Lambda^2)=\prod_{i=1}^{\tilde{d}}(1-l_{i}^2)\,.
	\end{aligned}
\end{equation*}
Thus, a sufficient condition to ensure $\Omega\in {\cal P}$ is to restrict $|l_i|<1$ for $i=1,\ldots,\tilde{d}$.

The matrix can be factorized as $\Omega=\Omega_L\Omega_L^\top$, where
\begin{equation*}\label{eq: Omega2Cholesky}
	\Omega_L=\begin{bmatrix}
		I_{d_1} & \vzero_{d_1 \times d_2} \\
		Q_2 \Lambda Q_1^\top & Q_2 \sqrt{I_{\tilde{d}} - \Lambda^2}
	\end{bmatrix}\,,
\end{equation*}
$\vzero_{d_1 \times d_2}$ is a $(d_1\times d_2)$ matrix of zeros, and $\Omega_L$ is a parsimonious (non-Cholesky) factor. 
To generate $\vz\sim N_d(\vzero,\Omega)$
first draw independent normals $\vepsilon=(\vepsilon_1^\top, \vepsilon_2^\top)^\top$ $\sim \mathcal{N}_{d_1 + d_2}(0, I)$, and then set
\begin{equation*}\label{eq: Omega2rep}
\vz = \Omega_{L} \vepsilon = \underline{\Phi}^{-1}(g_1(\vepsilon, \vlambda_{\mbox{\footnotesize vc}}))\,,
\end{equation*}
which also specifies the transformation $g_1$ with
$\vlambda_{\mbox{\footnotesize vc}} = (\vl^\top, \mbox{vec}(Q_1), \mbox{vec}(Q_2))^\top$, where $\vl = (l_1, \dots, l_{\tilde{d}})^\top$. The inverse of $\Omega$ has a fast to compute analytical expression;
%because $(\Omega/I_{d_1})^{-1} = Q_2\left( I_{\tilde{d}} - \Lambda^2 \right)^{-1} Q_2^\top$.
%	\begin{equation*}\label{eq: Omega2inv}
%		\Omega^{-1} =  \begin{bmatrix}
%			I_{d_1} + Q_1 \Lambda (I_{\tilde{d}} - \Lambda^2)^{-1} \Lambda Q_1^\top & -Q_1 \Lambda (I_{\tilde{d}} - \Lambda^2)^{-1} Q_2^\top \\
%			- Q_2 (I_{\tilde{d}} - \Lambda^2)^{-1}\Lambda Q_1^\top& Q_2 (I_{\tilde{d}} - \Lambda^2)^{-1} Q_2^\top
%		\end{bmatrix}\,,
%	\end{equation*}
see Part~B of the Online Appendix.

Because the Schur complement is only applicable to a $2 \times 2$ partitioned matrix, 
%adding more blocks in a recursive way will lead to non-identity matrix when using the Schur complement. Thus, 
it is difficult to extend this representation to more than $M = 2$ blocks. We label a GVC with this orthogonal pattern as ``GVC-O''. For some posteriors we also consider the special case where $Q_1=Q_2=I$ (so that only $\Lambda$ parameterizes the off-diagonal blocks at~\eqref{eq: Omega2}) and label this as ``GVC-I''.

%Substituting the re-parametrization tricks $g_1(\vepsilon, \vlambda_{\mbox{\footnotesize vc}})$ for GVC-F$p$ and GVC-O into equation~\eqref{eq: gvccdf} defines a distribution function $\Phi_d(\vz;\Omega)$ on $\vz$. As a result, the variational density $q(\vtheta)$ for GVC can be directly constructed by a change of variable from the Gaussian distribution.\footnote{We note that $\vz$ can be directly used in M1 and M2 because $\underline{\Phi}^{-1}(\vu_j)$ is a subvector of $\vz$, which is indicated by equation~\eqref{Omega1_rep} and~\eqref{eq: Omega2rep}.} The traditional re-parameterized gradients~\citep{ong2018gaussian} can be used to optimize ELBO:
%\begin{equation*}\label{eq: elbogradient1}
%	\nabla_{\vlambda} \fL (\vlambda) = \Exp_{f_\epsilon} \left\{ \left[ \frac{dg(\vepsilon, \vlambda)}{d \vlambda} \right]^\top \nabla_{\vtheta} \left( \log h(\vtheta) - \log q_\lambda(\vtheta)\right) \right\} \,.
%\end{equation*}

\subsection{Kendall vector copula}
A Kendall vector copula (KVC) is a special case of the hierarchical Kendall copula suggested by~\citet{brechmann2014hierarchical}. In a hierarchical Kendall copula, the within dependence of block $j$ is captured by a ``cluster copula'' $C_j$, and the dependence between different blocks is controlled by an $M$-dimensional ``nesting copula'' $C_0$. 
A general hierarchical Kendall copula is difficult to compute, but it
is tractable for the special case of a KVC 
where the cluster copulas are independence copulas (i.e. $C_j(\vu_j)= \prod_{s = 1}^{d_j}u_{j,s}$ for $j=1,\ldots,M$).
Let $c_0(\vv;\vlambda_{\mbox{\footnotesize vc}})=\frac{\partial^M}{\partial \vv}C_0(\vv;\vlambda_{\mbox{\footnotesize vc}})$ be the density of the nesting copula with $\vv=(v_1,\ldots,v_M)^\top$. Then, from 
\citet[p.87]{brechmann2014hierarchical}, the density of the $M$-block KVC is
\[c_v^{KV}(\vu_1, \dots, \vu_M;\vlambda_{\mbox{\footnotesize vc}}) =  c_0(\vv;\vlambda_{\mbox{\footnotesize vc}})\,,
\]
where $v_j = K_j(C_j(\vu_j))$ for $j = 1, \cdots, M$. Here, $K_j$ is the distribution function
of the random variable $V_j\equiv C_j(\vu_j)$, which is called the Kendall distribution function. For the KVC, $K_j(t)=\sum_{b = 0}^{d_j-1} \frac{t(-\ln t)^{b}}{b !}$
is derived in Appendix~\ref{app:kvc} and can be computed efficiently using Horner's method.

The between dependence of a KVC
can be specified using a latent vector $\vR = (R_1, \dots, R_M)^\top$ with the marginal 
distribution for each $R_j\geq 0$ given by an Erlang distribution with scale 1, shape $d_j$ and distribution function
\begin{equation*}\label{eq: Erlang}
	F_{R_j}(r_j) = 1-\sum_{b=0}^{d_j-1} \frac{r_j^b \exp(-r_j)}{b!}\,,\;\; r_j \in[0, \infty)\, .
\end{equation*}
This gives a generative procedure for the KVC with nesting copula $C_0$ outlined in
Algorithm~\ref{alg: sample from kvc} below, with a 
derivation given in Appendix~\ref{app:kvc}. 

Any existing parametric copula can be used for $C_0$, making the KVC an attractive as a VA. In our empirical work
we employ a Gaussian copula with correlation matrix $\Omega_0=\tilde{G}\tilde{G}^\top$ for $\tilde{G}=\mbox{diag}(GG^\top)^{-1/2}G$, with $G$ a lower 
triangular positive definite matrix. Thus, $\tilde{G}$ is a Cholesky factor, the leading 
diagonal elements of $G$ are unconstrained positive real numbers, 
and $\vlambda_{\mbox{\footnotesize vc}}=\mbox{vech}(G)$. This full rank parameterization
of $\Omega_0$ is
suitable for small to medium values of $M$ (typically $M<50$), although
other full rank~\citep{tan2018gaussian} or reduced rank~\citep{ong2018gaussian}
parameterizations for $\Omega_0$ may also be used.

%\begin{algorithm} [H]
%	\caption{Generation from an $M$-block KVC}
%	\label{alg: sample from kvc}
%	\begin{algorithmic}
%		\State Draw $\vv = (v_1, \cdots, v_M)^\top\sim C_0(\cdot;\vlambda_{\mbox{\footnotesize vc}})$, where $\vlambda_{\mbox{\footnotesize vc}}$ are the nesting copula parameters
%		\State
%		\For{j = 1 : $M$}
%		\State 1. Set $r_j = F_{R_j}^{-1}(1 - v_j)$.
%		\State 2. Draw $d_j$ independent samples $\ve_j=(e_{j,1},...,e_{j,d_j})^\top $ from $\text{Exp}(1)$.
%		\State 3. Set $\vs_j = \frac{\ve_j}{||\ve_j||}$, where $||\cdot||$ is the $L^1$ norm.
%		\State 4. Set $\vu_j = \exp{(-r_j \vs_j)}$. 
%		\EndFor
%		\State
%	\end{algorithmic}
%	Then $\vu = (\vu_1^\top, \cdots, \vu_M^\top)^\top\sim C_{v}^{KV}$ with nesting copula $C_0$ and parameters $\vlambda_{\mbox{\footnotesize vc}}$
%\end{algorithm}

An expression for the re-parameterization transformation $g_1$ is obtained directly from Algorithm~\ref{alg: sample from kvc} as
follows. Generate $\vepsilon_1\sim N(\vzero,I_M)$ and $d$ independent $e_{j,i}\sim \mbox{Exp}(1)$ random variables, then set $\vepsilon=(\vepsilon_1^\top,\ve_1^\top,\ldots,\ve_M^\top)^\top$ with $\ve_j=(e_{j,1},\ldots,e_{j,d_j})^\top$ for $j=1,\ldots,M$. The transformation is
\[
	\vu = g_1(\vepsilon,\vlambda_{\mbox{\footnotesize vc}}) \equiv \exp\left\{-\vr.\mbox{diag}\left(\frac{\ve_1}{||\ve_1||},\ldots,\frac{\ve_M}{||\ve_M||}\right)\right\}\,, \\
\]
where $\vr=\left(F_{R_1}^{-1}(1-v_1)\viota_{d_1},\ldots,F_{R_M}^{-1}(1-v_M)\viota_{d_M}\right)$ is a $d$ dimension row vector, $\viota_{d_j}$ is a $d_j$ dimension row vector of ones, $\vv=\underline{\Phi}(\tilde{G}\vepsilon_1)$, ``diag($\vx$)''
denotes a diagonal matrix with leading diagonal elements $\vx$, and the exponential function is applied element-wise. We label
this Kendall vector copula with a Gaussian nesting copula as ``KVC-G''. Finally, we note that
using a Gaussian copula for $C_0$ simplifies the re-parameterized gradients at~\eqref{eq: elbogradient} as discussed further in Part C of the Online Appendix.
\begin{algorithm} [H]
	\caption{Generation from an $M$-block KVC}
	\label{alg: sample from kvc}
	\begin{algorithmic}
		\State Draw $\vv = (v_1, \cdots, v_M)^\top\sim C_0(\cdot;\vlambda_{\mbox{\footnotesize vc}})$, where $\vlambda_{\mbox{\footnotesize vc}}$ are the nesting copula parameters
		\State
		\For{j = 1 : $M$}
		\State 1. Set $r_j = F_{R_j}^{-1}(1 - v_j)$.
		\State 2. Draw $d_j$ independent samples $\ve_j=(e_{j,1},...,e_{j,d_j})^\top $ from $\text{Exp}(1)$.
		\State 3. Set $\vs_j = \frac{\ve_j}{||\ve_j||}$, where $||\cdot||$ is the $L^1$ norm.
		\State 4. Set $\vu_j = \exp{(-r_j \vs_j)}$. 
		\EndFor
		\State
	\end{algorithmic}
	Then $\vu = (\vu_1^\top, \cdots, \vu_M^\top)^\top\sim C_{v}^{KV}$ with nesting copula $C_0$ and parameters $\vlambda_{\mbox{\footnotesize vc}}$
\end{algorithm}

\subsection{Discussion of vector copulas}

\begin{table}[htbp]
	\begin{center}
		\caption{Characteristics of Different Vector Copulas \& Parameterizations}
		\label{tab: vector_copulas}
		\begin{tabular}{llll}
			\toprule
			\toprule
			Vector Copula & $M$ & Parameters & $\mbox{dim}(\vlambda_{\mbox{\footnotesize vc}} )$ \\
			\midrule
			GVC-F$p$ & Unconstrained & $\zeta,B$ &$pd + 1$ \\
			GVC-O &2 &$Q_1,Q_2,\Lambda$ & $(d_1+d_2+1)\tilde{d}$ \\
			GVC-I & 2 &$\Lambda$ &$\tilde{d}$ \\
			KVC-G & Unconstrained &$G$ &$M(M+1)/2$ \\
			\bottomrule
			\bottomrule
		\end{tabular}
	\end{center}
	Note: $p$ is the number of factors used in the GVC with factor patterned $\Omega$; $d_1$ and $d_2$ are the number of rows in $Q_1$ and $Q_2$ in the GVC with orthogonal patterned $\Omega$; and $\tilde{d} = \min{(d_1,d_2)}$.
\end{table}

Table~\ref{tab: vector_copulas} summarizes characteristics of the vector copulas
discussed above, and we make some comments on their suitability for capturing dependence between blocks in different posteriors.
First, for GVC-F$p$ the number of variational parameters $\mbox{dim}(\vlambda_{\mbox{\footnotesize vc}})$ does not vary with the number of blocks $M$, so that this 
vector copula is attractive for partitions with large $M$. 
Second, the number of parameters of a KVC-G is determined by $M$ and not $d$. In addition, blocks 
with large $d_j$ incur minimal computational burden, making the KVC-G an attractive choice for 
high $d$, $M$ and big blocks with large $d_j$. However, with a KVC, 
dependence between blocks is captured by the $M$-dimensional nesting copula $C_0$, which
can prove limiting for some posteriors. For example,
the KVC-G  has only a single scalar parameter to capture dependence between each pair of blocks.
Third, both GVC-O and GVC-I are restricted to two blocks. However, as we show in two examples,
they are attractive to capture posterior dependence for popular global-local shrinkage priors. Moreover, they can be used as a marginal of other vector copula model when $M>2$. Fourth,
the factor patterned correlation matrix is not a unique parameterization. But this is not a problem, as over-identified fixed forms can be effective approximations; see the discussion in \cite{ong2018gaussian}.

Finally, 
 any existing copula density with multivariate marginals that are independence copulas, are well-defined vector copulas. Therefore, the identification of additional 
vector copulas that prove suitable as VAs is a promising line of research.

\section{Applications} \label{sec4}
To show the versatility of our method we apply it to estimate four different statistical models using 16 datasets, each having complex posteriors.
The first two models use global-local shrinkage priors, which result in posteriors
that are approximated well using GVC-I.    
The second two models have correlated
latent variables with a posterior for which GVC-F$p$ and KVC-G are suitable approximations. 
For comparison, we use the following (non-vector copula model) VAs as benchmarks:
\begin{itemize}
	\item GMF: Fully factorized Gaussian mean field with $d$ independent Gaussians.
	\item G-F$p$: Gaussian with a factor covariance structure with $p$ factors as in~\cite{ong2018gaussian}.
	\item GC-F$p$: Conventional Gaussian copula constructed by element-wise YJ transformations and a factor correlation matrix with $p$ factors as in~\cite{smith2020high}.
	\item BLK: $M$ independent blocks of either M1 or M2 ($w=1$) multivariate Gaussian marginals (so that $\eta_{j,i}=1$ for all $i,j$) .
	\item BLK-C: As with BLK, except that $\veta_1,\ldots,\veta_M$ are learned (so that each multivariate marginal is a Gaussian copula).  
\end{itemize}
The SGD algorithm is used to calibrate all VAs. To ensure 
accurate results 40,000 steps are used, although the optimum is often reached 
within 5,000 steps in our examples.
The VAs GC-F$p$ and BLK-C have skewed univariate marginals, while G-F$p$ and GC-F$p$
capture dependence in $\vtheta$ using $p$ common sources of variation across all elements, and are among the best performing black-box VI methods that balance accuracy and speed for high $d$.
Accuracy of a VA is measured using the ELBO at~\eqref{eq: elboq}, computed as its median value over the
final 1000 steps of the SGD algorithm. A
higher ELBO value corresponds to a lower KLD between the VA and the target posterior.

We note that the first two models are difficult to estimate using MCMC for large datasets, and
we show that VCVI is a fast and feasible alternative. The last two models can be estimated readily using MCMC, which allows the accuracy of VCVI to be demonstrated. The dimensions of the posteriors
in our applications are listed in Table~A2 in Part~D of the
Online Appendix.

\subsection{Example 1: Sparse logistic regression with horseshoe regularization}\label{sec:eg1}
The first application is a sparse logistic regression, where there is a large number of 
covariates $m$ relative to the number of observations $n$. In this case it is usual to regularize
the coefficient vector $\vbeta = (\beta_1, \dots, \beta_m)^\top$, and here we use the horseshoe prior~\citep{carvalho2010horseshoe} given by
\begin{equation*}
		\beta_i | \xi, \delta_i \sim \mathcal{N} (0, \delta_i^2 \xi^2)\,,\;\;
		\delta_i \sim \mathcal{C}^+(0,1)\,,\;\;
		\xi \sim \mathcal{C}^+ (0,1)\,,
\end{equation*}
where $\fC^+$ denotes the half-Cauchy distribution, $\delta_i$ is a local shrinkage parameter and $\xi$ is the global shrinkage parameter. The posterior has a funnel-shaped posterior~\citep{ghosh2018structured} that is hard to approximate and to simplify the geometry we use a non-centered parameterization~\citep{ingraham2017variational} that sets $\beta_i = \alpha_i \delta_i \xi$ with prior $\alpha_i \sim \fN(0,1)$. The model parameters are  $\vtheta = (\valpha^\top, \tilde{\vdelta}^\top, \tilde{\xi})^\top$
where $\valpha = (\alpha_1, \dots, \alpha_m)^\top$, $\tilde{\vdelta} = (\tilde{\delta}_1, \cdots, \tilde{\delta}_m )^\top$, $\tilde{\xi} = \log(\xi)$ and $\tilde{\delta_i} = \log(\delta_i)$. 

We choose four lower dimensional datasets from~\citet{ong2018gaussian}, named as \textit{krkp}, \textit{spam}, \textit{iono}, \textit{mushroom}. For the high-dimensional situation, we choose the dataset \textit{cancer} from~\citet{ong2018gaussian} and the 
{\em QSAR Oral Toxicity} dataset from the UCI Machine Learning Repository. 
% \href{https://archive.ics.uci.edu/dataset/508/qsar+oral+toxicity}{\textit{QSAR oral toxicity}} from UCI Machine Learning Repository. 
Additionally, we use the function
``make\_classification''  %\href{https://scikit-learn.org/1.5/modules/generated/sklearn.datasets.make_classification.html}{\texttt{make\char`_classification}} 
from Scipy to simulate three datasets with 5, 20 and 100,000 observations, each with 10,000 covariates (with $\beta_i\neq 0$ for 2,000). The top rows of Table~\ref{tab: logitreg_ELBOs} lists these nine datasets and their size.

Two likely features of the posterior that guide our VA selection 
are that (i)~the pairs $(\alpha_i,\tilde{\delta}_i)$ will be dependent, and (ii)~when the signal is sparse (i.e. when the ratio $n/m$ is low as in our datasets) within dependence in both $\valpha$ and $\tilde{\vdelta}$ is weak.
 We partition $\vtheta$ into $(\valpha^\top,\tilde{\vdelta}^\top,\tilde \xi)$ and consider
the following 3-block vector copula models as VAs: 
\begin{itemize}
	\item A1/A2: GVC-F5 with marginals M1 $(L_j = I)$ for all three blocks	
	\item A3/A4: Bivariate GVC-I with marginals M1 $(L_j = I)$ for $\valpha$ and $\tilde{\vdelta}$, and independent $\tilde \xi$
	\item A5/A6: Bivariate GVC-I with marginals M2 $(w=1)$ for $\valpha$ and $\tilde{\vdelta}$, and independent $\tilde \xi$
%	\item A7: Nested vector copula with Bivariate GVC-I for $(\valpha,\tilde{\vdelta})$ with marginals M1 $(L_j = I)$ which is linked to $\xi$ by a bivariate KVC-G. 
\end{itemize}
%The between structures A1-A7 are summarized in Figure~\ref{fig: logitreg_VCVI_types}. 
For approximations A1, A3 and A5 we set $k_{\veta_j}$ to the identity transformation (i.e. where $\eta=1$ in~\eqref{eq: IYJ}) resulting in Gaussian marginals. For approximations A2, A4 and A6, $\veta_1,\veta_2,\veta_3$ are learned, so that the two multivariate marginals for $\valpha$ and $\tilde{\vdelta}$ are conventional Gaussian copula models, and all univariate marginals are potentially skewed. The benchmarks BLK/BLK-C have the same multivariate M2 marginals as A5/A6, but
assume independence between all three blocks.

%\begin{figure}[htbp]
%	\centering
%	\begin{subfigure}[b]{0.28\textwidth}
%		\centering
%		\includegraphics[width=\textwidth]{PLOT/(a).pdf}
%	\end{subfigure}
%	\hfill
%	\begin{subfigure}[b]{0.28\textwidth}
%		\centering
%		\includegraphics[width=\textwidth]{PLOT/(b).pdf}
%	\end{subfigure}
%	\hfill
%	\begin{subfigure}[b]{0.28\textwidth}
%		\centering
%		\includegraphics[width=\textwidth]{PLOT/(c).pdf}
%	\end{subfigure}
%	
%	\caption{Different types of VCVI for the logistic regression with a horseshoe prior.}
%	\label{fig: logitreg_VCVI_types}
%\end{figure}

\begin{sidewaystable}[p]
	\begin{center}
		\caption{ELBO values for different VAs (rows) in the regularized logistic regression with nine datasets (columns)}
		\label{tab: logitreg_ELBOs}
		\begin{tabular}{lccccccccc}
			\toprule
			\toprule
			Dataset& \multicolumn{4}{c}{Low-Dimensional Real} & \multicolumn{2}{c}{High-dimensional Real} & \multicolumn{3}{c}{High-Dimensional Simulated}\\
			\cmidrule(lr){2-5} \cmidrule(lr){6-7} \cmidrule(lr){8-10}
			Name & krkp & mushroom & spam & iono & cancer & qsar & simu1 & simu2 & simu3 \\
			Size ($n$x$m$) & 3196x38 & 8124x96 & 4601x105 & 351x112 & 42x2001 & 8992x1024 & 5Kx10K & 20Kx10K & 100Kx10K \\
			\midrule
			\multicolumn{10}{l}{\underline{\textit{Benchmark VAs}}:} \\
			GMF & -382.01 & -120.85 & -856.77 & -98.82 & \textbf{-77.59} & -2245.10 & -3641.64 & -10844.22 & -40735.67 \\
			\cdashline{2-10} \\
			G-F5 & 8.80 & 4.52 & 5.93 & 2.50 & -0.53 & -0.53 & -6.16 & -18.47 & -38.21 \\
			GC-F5 & 10.92 & 14.21 & 15.43 & 7.07 & -0.66 & 40.94 & 84.75 & 198.37 & 242.80 \\
			G-F20 & 19.74 & 2.55 & 13.78 & 3.21 & -1.92 & -15.19 & -25.71 & -70.41 & -145.87 \\
			GC-F20 & 20.87 & 8.86 & 24.84 & 8.45 & -2.02 & 22.30 & 65.31 & 153.10 & 129.14 \\
			BLK & -1.41 & 0.42 & -2.08 & -3.76 & -2.81 & 2.74 & -38.34 & -44.31 & -73.10 \\
			BLK-C & -8.86 & 10.02 & -0.09 & 5.32 & -0.33 & 35.16 & -12.93 & 212.23 & 263.74 \\
			\midrule
			\multicolumn{10}{l}{\underline{\textit{Vector Copula Model VAs}}:} \\
			A1: (GVC-F5 \& M1) & 15.43 & 4.90 & 10.07 & 1.81 & -0.55 & 17.84 & CI & CI & CI \\
			A2: (GVC-F5 \& M1-YJ) & 19.57 & 15.24 & 19.33 & 8.63 & -0.61 & 56.13 & CI & CI & CI \\
			\cdashline{2-10} \\
			A3: (GVC-I \& M1) & 25.57 & 10.50 & 25.96 & 4.77 & -0.31 & 47.72 & 51.00 & 409.67 & 1229.27 \\
			A4: (GVC-I \& M1-YJ) & \textbf{36.98} & \textbf{21.09} & \textbf{42.11} & 10.26 & -0.27 & \textbf{87.93} & \textbf{185.63} & \textbf{753.43} & \textbf{1929.11} \\
			\cdashline{2-10} \\
			A5:  (GVC-I \& M2) & 25.00 & 10.29 & 25.95 & 4.94 & -0.49 & 45.89 & 19.34 & 362.90 & 1172.71 \\
			A6: (GVC-I \& M2-YJ) & 33.01 & 20.05 & 40.25 & \textbf{10.55} & -1.75 & 85.45 & 121.51 & 700.10 & 1857.95 \\
			\bottomrule
			\bottomrule
		\end{tabular}
	\end{center}
	Note: ELBO values are reported for GMF, and the differences from these values are reported for the other VAs. Higher values correspond to higher accuracy, and the bold value in each column indicates the highest ELBO value for the dataset. %The values reported are the median of ELBO values over the last 1000 steps of the SGD algorithm.
	 ``CI'' represents computationally infeasible using our Python code. Detailed descriptions of the different VAs are provided in the text.	
\end{sidewaystable}

\begin{figure}[htbp]
	\centering
	\includegraphics[width=0.8\textwidth]{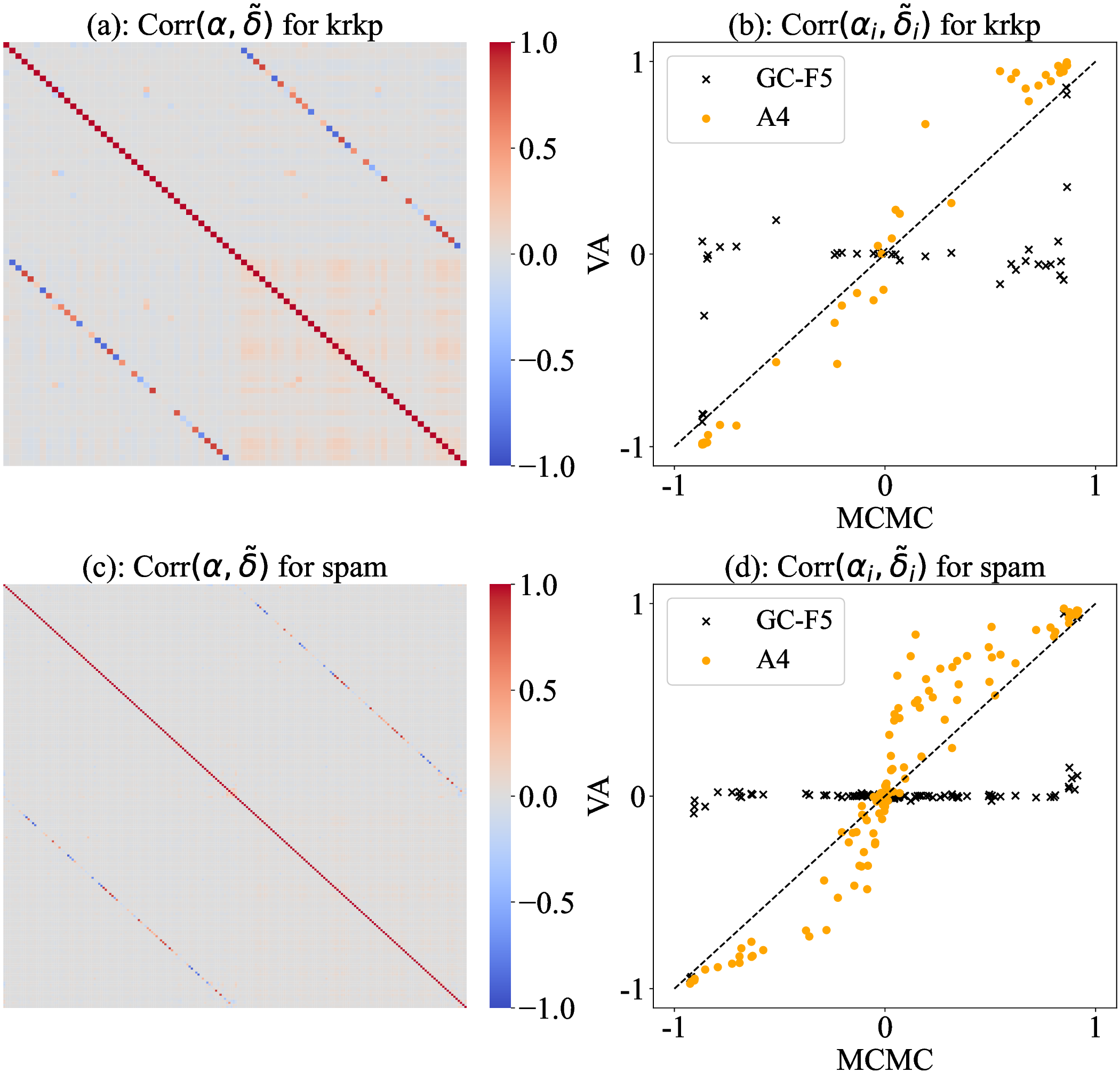}
	\caption{Posterior summaries for {\em krkp} (top row) and {\em spam} (bottom row). Panels~(a,c) are heatmaps of the matrix of Spearman correlations for $(\valpha^\top, \tilde{\vdelta}^\top)$ computed exactly using (slow) MCMC. Panels~(b,d) are scatterplots of $\{\mbox{corr}(\alpha_i,\tilde{\delta}_i); i=1,\ldots,m\}$, with exact posterior values on the horizontal axis, and variational approximations on the vertical axis for GC-F5 (black crosses) and A4 (orange dots).}
	\label{fig: logitreg_heatmap_corr}
\end{figure}

Table~\ref{tab: logitreg_ELBOs} reports the ELBO value (computed as the median ELBO over the last 1000 steps of the SGD algorithm) for the GMF benchmark, and the difference between this and
the values for the other VAs. We make five observations. First, allowing for dependence in the posterior in any way increases accuracy compared to GMF for all examples except the {\em cancer} dataset. This dataset is extreme, with only $n=42$ observations on $m=2001$ covariates, so that the prior dominates the likelihood in the posterior.
Second, the popular G-F$p$ and GC-F$p$ (with either $p=5$ or $20$ factors) 
benchmark VAs are less accurate than the proposed vector copula models for most datasets. 
Third, allowing for skewed univariate marginals by learning the YJ transformation parameters (as in GC-F$p$, BLK-C, A2, A4 and A6) increases accuracy in almost all cases. 
Fourth, the vector copula models A4 and A6 are the most accurate VAs for all datasets, except {\em cancer}. Fifth---and an important finding of this study---is that accounting for dependence between blocks improves accuracy. This can be seen with A5/A6 dominating BLK/BLK-C, which are benchmark VAs with the same multivariate marginals but where the three blocks are independent.

Figure~\ref{fig: logitreg_heatmap_corr} shows why GVC-I performs well. 
Panels~(a,c) plot the posterior Spearman correlation matrices of $(\valpha^\top, \tilde{\vdelta}^\top)$ computed exactly using MCMC for {\em krkp} and {\em spam}. The main feature is that the pairs $(\alpha_i,\tilde{\delta}_j)$ exhibit
high correlation when $i=j$ (but not when $i\neq j$), which is common when using global-local shrinkage priors. This feature is captured parsimoniously by $\Lambda$ in GVC-I. 
Panels~(c,d) plot the exact posterior pairwise correlations of pairs $(\alpha_i,\tilde{\delta}_i)$ against their variational estimates for GC-F5 and A4. The low rank GC-F5 cannot approximate these values well, whereas A4 is much better in doing so. Ignoring them (as in GMF, BLK and BLK-C) also reduces accuracy.

\begin{figure}[htbp]
	\centering
	\includegraphics[width=\textwidth]{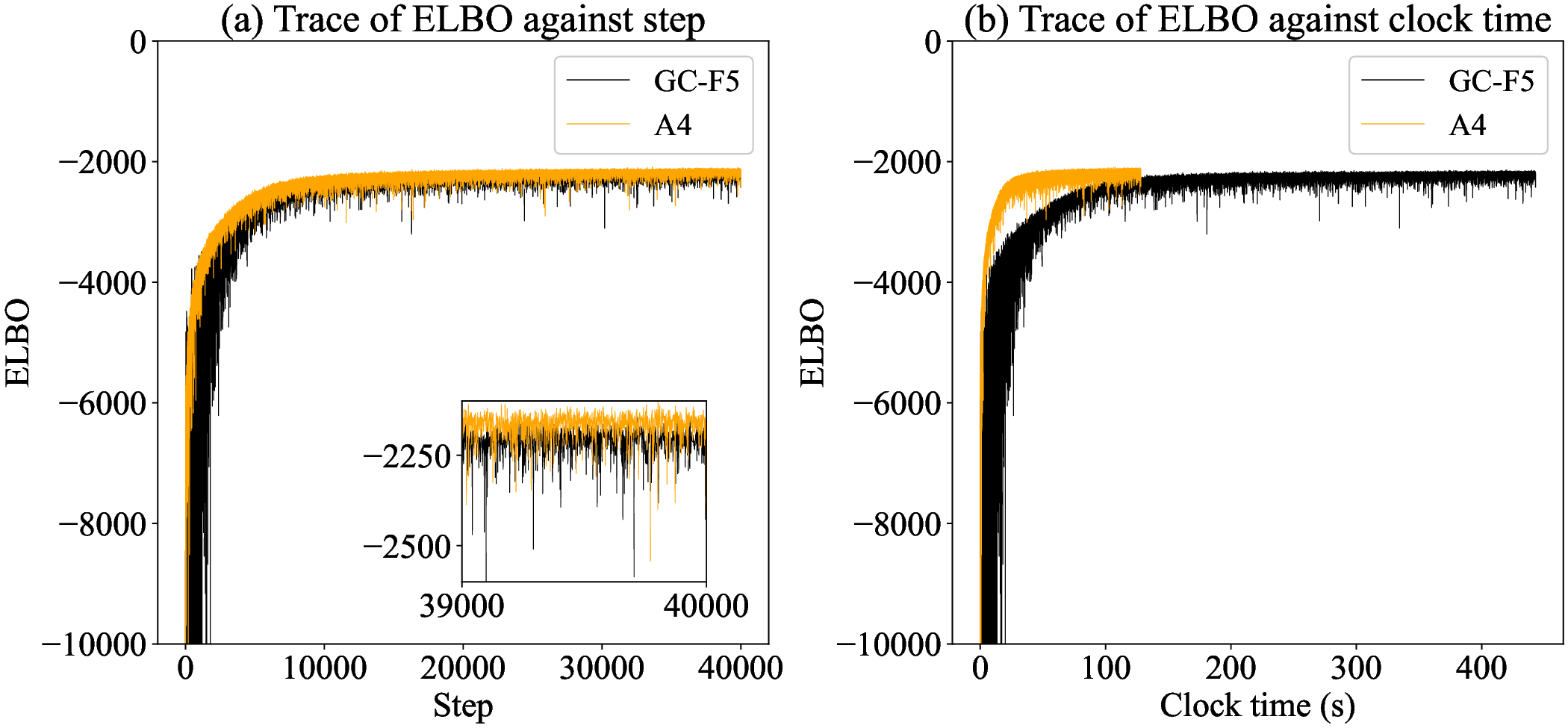}
	\caption{Plot of the ELBO values for the dataset {\em qsar} ($n=8992, m=1024$) for GC-F5 (black line) and A4 (orange line) against (a) SGD step number, (b) wall clock time.}
	\label{fig: logitreg_trace_qsar}
\end{figure}

The proposed vector copula model VAs can also be fast to calibrate. Table~A1 in Part~D of the Online Appendix provides computation times
of the SGD algorithm implemented in Python. 
The VAs A3/A4 are almost as fast as GMF, while A5/A6 are only slightly slower than BLK/BLK-C which share the same marginals. 
The VAs A1/A2 are slow as $d$ increases (becoming infeasible for $d=10,000$ in the simulated datasets). This is because a fast implementation of the $O(pd^2)$ update of the Cholesky factor $A$ in Section~\ref{sec:GVCFp} was unavailable in Python, and we employed full $O(d^3)$
evaluation instead. Figure~\ref{fig: logitreg_trace_qsar} plots ELBO values for A4 and GC-F5 against (a)~SGD step, (b)~wall clock time. Not only is A4 much faster per step, it converges in fewer steps to a higher value.

In a final comment, $\tilde\xi$ is treated as independent in the vector copula model VAs. We have also experimented with a nested $M=3$ vector copula, where the GVC-I for $(\valpha,\tilde\vdelta)$ is the marginal of a bivariate KVC-G that also includes $\tilde\xi$. This nested vector copula was slightly more accurate than A4 for most datasets;
see Part~E of the Online Appendix for further details. 

\subsection{Example 2: Correlation matrix with LASSO regularization}\label{sec:correg}
The second application of VCVI is to estimate a regularized correlation matrix of a Gaussian copula model.\footnote{This copula model is for the observed data, and is not to be confused with the vector and conventional copula
	models that are used as VAs of the posterior distribution.} We use this model to estimate inter-state dependence 
in annual changes in wealth inequality (measured by the Gini coefficient) for $r$ U.S. states 
using panel data 
from 1916 to 2018 provided by Mark Frank on his website 
{\em www.shsu.edu/eco\_mwf/inequality.html}. For each state, the observations are transformed
using nonparametric marginals followed by the standard  normal quantile function.
The $(r\times r)$ correlation matrix $\Sigma$ of these transformed data is the parameter of the Gaussian copula.
\cite{pinheiro1996unconstrained} show that the lower triangular Cholesky factor of $\Sigma$
can be uniquely parameterized by angles $\tilde{\vbeta} = (\tilde{\beta}_1, \dots, \tilde{\beta}_{r(r-1)/2})^\top $.
 Each angle $\tilde{\beta}_i\in [0, \pi)$ is transformed to the real line by setting $\beta_i = \Phi^{-1}( \tilde{\beta}_i / \pi )$; see~\citet{smith2023implicit} for a detailed description of this parameterization and example.
 
Regularization is useful because there are only $n=103$ observations to estimate the $r(r-1)/2$ parameters. To illustrate the flexibility of our VCVI method we employ the Bayesian LASSO of~\cite{figueiredo2003adaptive}. This is a global-local shrinkage prior where
\begin{equation*}
		\beta_i | \delta_i \sim \mathcal{N} (0, \delta_i)\,\;\; \mbox{and}\;\; \delta_i | g \sim \mbox{Exp}(g/2)\,,
\end{equation*}
with $\mbox{Exp}(\kappa)$ denoting an exponential distribution with rate $\kappa$. 
It can be shown that a partial correlation is zero iff $\beta_i=0$ \citep{lewandowski2009} so that shrinking 
$\beta_i$ towards zero is equivalent to shrinking these partial correlations towards zero too.
%AP: IN THE SENTENCE BELOW THERE IS A LOT OF REPETITION
To simplify the geometry of the posterior, a non-centered parameterization for $\beta_i$ is used, such that $\beta_i = \alpha_i \sqrt{\delta_i}$ with $\alpha_i\sim N(0,1)$ and $\sqrt{g} \sim \mbox{Exp}(1)$. 
Let $\tilde{\delta}_i = \log(\delta_i)$, $\tilde{g} = \log(g)$, $\valpha = (\alpha_1, \dots, \alpha_{r(r-1)/2})^\top$ and $\tilde{\vdelta} = (\tilde{\delta}_1, \dots, \tilde{\delta}_{r(r-1)/2})^\top$, then $\vtheta = (\valpha^\top, \tilde{\vdelta}^\top, \tilde{g})^\top$. 
%We use the same U.S. income inequality data as in~\citet{smith2023implicit}, which contains the annual Gini coefficient from 1916 to 2018 for 49 states. Observations of the time series are set as $y_{t,j} = GINI_{t,j} - GINI_{t-1,j}$, where $GINI_{t,j}$ is the GINI coefficient for state $j$ at time $t$.

The same VAs and blocking strategies are adopted as with the previous example because the between dependence in $\vtheta$ is similar. There are 49 U.S. states in our data (Alaska and Hawaii are omitted due to missing data, and DC is included as a state), which we order by population 
and compute inference for $\Sigma$ when $r=5, 10, 20, 30, 49$ states (so that there five datasets). 
The accuracy of the VAs are summarized in Table~\ref{tab: corrmatrix_ELBOs}.
The same five observations made in the previous example apply equally here to this 
different statistical model and global-local shrinkage prior. 
\begin{table}[htbp]
	\begin{center}
		\caption{ELBO values for different VAs for the regularized correlation matrix $\Sigma$}
		\label{tab: corrmatrix_ELBOs}	
		\begin{tabular}{lccccc}
			\toprule
			\toprule
			Number of U.S. States & $r=5$ & $r=10$ &$r=20$ &$r=30$ &$r=49$ \\
			Dimension of $\vtheta$ & 21 & 91 & 381 & 871 & 2353 \\
			\midrule
			\multicolumn{6}{l}{\underline{\textit{Benchmark VAs}}:} \\
			GMF & -553.93 & -1039.24 & -2038.91 & -3152.38 & -5548.02 \\
			\cdashline{2-6} \\
			G-F5 & 5.71 & 6.10 & 4.53 & 2.87 & DNC \\
			GC-F5 & 6.42 & 10.29 & 12.21 & 20.83 & DNC \\
			BLK & 0.21 & 3.16 & 1.03 & 0.14 & -1.36 \\
			BLK-C & 0.66 & 5.72 & 13.14 & 32.76 & 89.94 \\
			\midrule
			\multicolumn{6}{l}{\underline{\textit{Vector Copula Model VAs}}:} \\
			A1: (GVC-F5 \& M1) & 5.94 & 8.61 & 6.42 & 6.83 & 5.16 \\
			A2: (GVC-F5 \& M1-YJ) & 8.01 & 11.97 & 19.05 & 39.09 & 97.66 \\
			\cdashline{2-6} 
			A3: (GVC-I \& M1) & 8.51 & 28.54 & 53.26 & 75.52 & 94.91 \\
			A4: (GVC-I \& M1-YJ) & \textbf{10.91} & \textbf{35.48} & \textbf{71.07} & 109.90 & \textbf{181.98} \\
			\cdashline{2-6} 
			A5:  (GVC-I \& M2)  & 8.59 & 28.61 & 53.25 & 75.47 & 94.29 \\
			A6: (GVC-I \& M2-YJ) & 10.64 & 34.96 & 70.96 & \textbf{110.06} & 181.25 \\
			%			\cdashline{2-6} 
			%			A7 & \textbf{10.93} & 35.34 & \textbf{71.17} & 109.95 & \textbf{182.03} \\
			\bottomrule
			\bottomrule
		\end{tabular}
	\end{center}
Note: The posterior is for the $(r\times r)$ regularized correlation matrix $\Sigma$ of the Gaussian copula model for U.S. wealth inequality panel data. The columns give results for $r=5$, 10, 20, 30 and 49 U.S. states, and the VAs (rows) are described
in Section~\ref{sec:correg}. ELBO values are reported for GMF, and the differences from these values are reported for the other VAs. Higher values correspond to greater accuracy, with the largest value
in each column in bold. 
%The values reported are the median of ELBO values over the last 1000 steps of the SGD algorithm. 
DNC denotes ``Did Not Converge' within 200,000 iterations.
 %Note: The models for 5 ,10, 20, 30 and 49 states are trained by 50,50,100,150 and 200 thousand steps using Adam, respectively. For each VA, the median of ELBO values over the last 1000 steps is compared to that from GMF, and the improvement (if positive) is shown in the table. A higher value corresponds to a more accurate VA, and bold value indicates the optimal VA for a dataset. DNC denotes ``Did Not Converge' within 200 thousand iterations. We use the median here because ELBO values are highly skewed.
\end{table}

\begin{figure}[htbp]
	\centering
	\includegraphics[width=\textwidth]{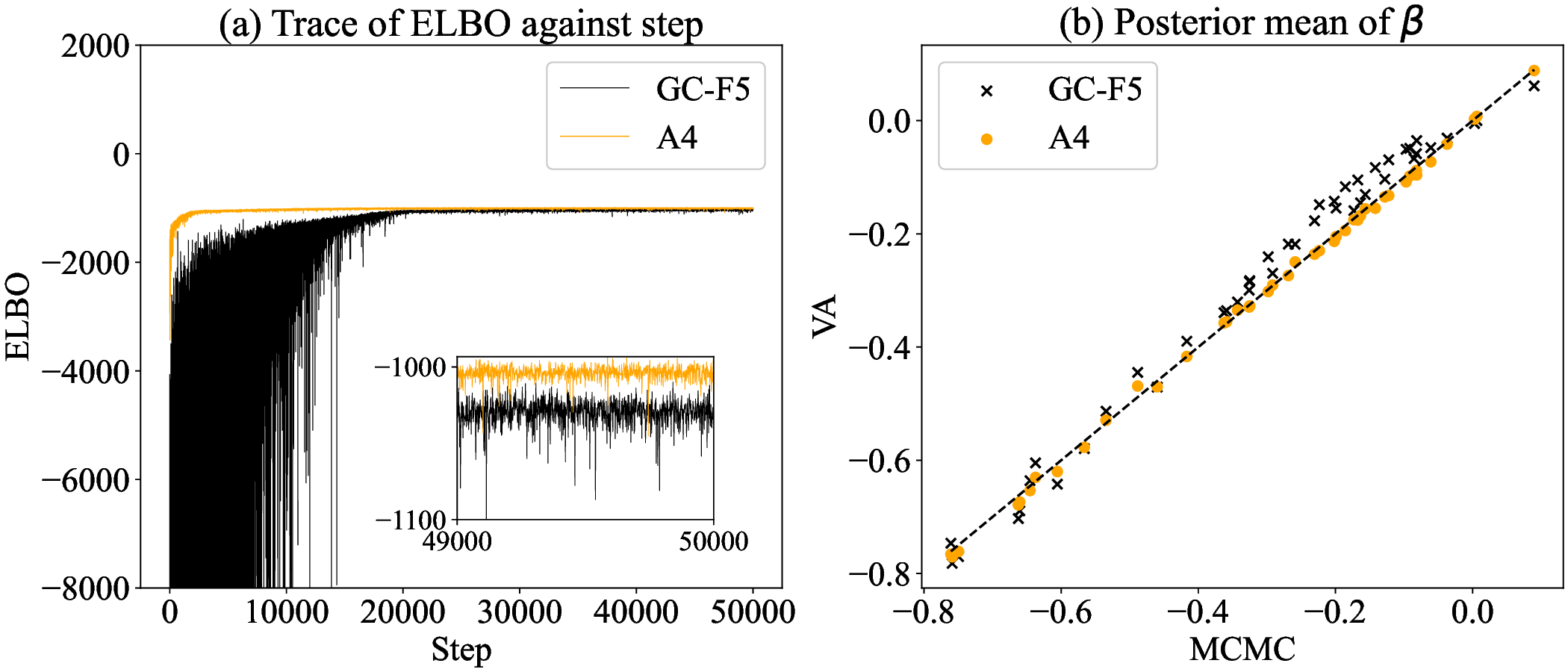}
	\caption{Comparison of two approximations GC-F5 (black) and A4 (orange) for the regularized correlation matrix example with $r=10$. Panel~(a) plots the ELBO values against SGD step. Panel~(b) plot the exact posterior mean of $\vbeta$ computed using MCMC (horizontal axis) against the variational posterior means (vertical axis). Greater alignment on the 45 degree line corresponds to higher accuracy.}
	\label{fig: corr_10states}
\end{figure}

In addition, we found
that SGD was slow to converge for G-F$p$ and CG-F$p$ models, whereas the vector copula models were fast. To illustrate, 
Figure~\ref{fig: corr_10states} show the performance of VAs GC-F5 and A4 on the $r=10$ state dataset. Panel~(a) shows that the A4 was not only the most accurate VA, but that the SGD algorithm
obtained its optimum within a few thousands steps, whereas it took 20,000 steps for GC-F5.
For the case of $r=49$ states, G-F5 and GC-F5 did not converge in 200,000 steps.  
When $r=10$ it is possible to compute the exact posterior using (slow) MCMC. Figure~\ref{fig: corr_10states}(b) plots the true posterior mean of $\vbeta$ against the variational posterior means for the two VAs, further demonstrating the increased accuracy of the VCVI method. %Finally, we note that in experiments we find that the vector copula model VAs dominate the factor models also
%when the horseshoe prior is used.

\subsection{Example 3: Unobserved Component Stochastic Volatility Model}\label{sec:egucsv}
\cite{chan2017stochastic} notes that a successful univariate model for inflation is a stochastic volatility model with
an unobserved component (UCSV). We apply this to $T=282$  
quarterly U.S. inflation observations $y_t =\log(P_t) - \log(P_{t-1})$, where $P_t$ is the seasonally-adjusted GDP price deflator from 1954Q1 to 2024Q3.\footnote{The implicit price deflator series [GDPDEF], was retrieved from the Federal Reserve Bank of St. Louis; https://fred.stlouisfed.org/series/GDPDEF, January 16, 2025.} The UCSV model decomposes $y_t =\mu_t+e^{\zeta_t/2}\varepsilon_t$, with disturbance $\varepsilon_t\sim \mathcal{N}(0,1)$ and first
order autoregressive processes in the mean and logarithmic volatility 
\begin{equation*}%\label{SVUC}
	\begin{aligned}
	\mu_t|\mu_{t-1},\bar{\mu},\rho_{\mu},\sigma_{\mu} &\sim \mathcal{N}(\bar{\mu} + \rho_{\mu}(\mu_{t-1} - \bar{\mu}), \sigma_{\mu}^2)\,, \\
	\zeta_t|\zeta_{t-1},\bar{\zeta},\rho_{\zeta},\sigma_{\zeta} &\sim \mathcal{N}(\bar{\zeta} + \rho_{\zeta}(\zeta_{t-1} - \bar{\zeta}), \sigma_{\zeta}^2)\,,
	\end{aligned}
\end{equation*}
with $\mu_0 \sim \fN(\bar{\mu}, \sigma^2_{\mu}/(1 - \rho^2_{\mu}))$ and $\zeta_0 \sim \fN(\bar{\zeta}, \sigma^2_{\zeta}/(1 - \rho^2_{\zeta}))$. As is common for this model, proper priors
are adopted that bound $0<\rho_{\mu}<0.985$ and $0<\rho_{\rho}<0.985$, so that the mean and log-volatility
processes are stationary, numerically stable and have positive serial dependence.\footnote{The priors are $\bar{\mu} \sim \fN(0.5, 1000)$, $\rho_{\mu} \sim \fN(0.9,0.04)$ with $0<\rho_{\mu}<0.985$, $\sigma^2_{\mu} \sim \mathcal{IG}(2, 0.001)$, $\bar{\zeta} \sim \fN(-2, 1000)$, $\rho_{\zeta} \sim \fN(0.9,0.04)$ with $0<\rho_{\zeta}<0.985$ and $\sigma^2_{\zeta} \sim \mathcal{IG}(2,0.001)$.}
 
\begin{table}[tbp]
	\begin{center}
		\caption{Size, speed and accuracy of VAs in the UCSV example}
		\label{tab: SVUC_ELBOs}
		\resizebox{1.0\textwidth}{!}{
			\begin{tabular}{lccc}
				\toprule
				\toprule
				& \# of Variational Parameters & Time (s) Per 1000 Steps &  ELBO \\
				\midrule
				\underline{\textit{Benchmark VAs}}: \\
				GMF & 1144 & 0.17 & -95.26 \\
				GC-F5 & 4566 & 1.48 & -92.99 \\
				GC-F20 & 12966 & 1.60 & -97.78 \\
				BLK-C & 2295 & 3.04 & -59.75 \\
				\midrule
				\underline{\textit{Vector Copula Model VAs}}: \\
				GVC-F5  & 5146 & 10.01 & -58.91 \\
				GVC-F20  & 13546 & 10.08 & \textbf{-57.79} \\
				GVC-I & 2578 & 3.01 & -58.90 \\
				KVC-G & 2301 & 3.60 & -59.71 \\
				\bottomrule
				\bottomrule
			\end{tabular}
		}
	\end{center}
%	Note: All methods are trained by Adam by 30,000 steps.
	Note: The approximations BLK-C, GVC-F$p$, GVC-I and KVC-G all have the same M1 marginals with band one $L_j^{-1}$ matrices for $\vmu$ and $\vzeta$ and unconstrained lower triangular matrix $L_3$ for $\vtheta_3$. The number of variational parameters is $|\vlambda|$. Times are reported for implementation in Python on an Apple M3 Max laptop. %The ELBO values are the median from the last 1000 steps of the SGD algorithm, with higher values indicating greater accuracy.  
	Higher ELBO values indicate greater accuracy, with the largest value in bold. 
\end{table}

This model is readily estimated using MCMC~\citep{chan2017stochastic} so that the accuracy of 
the variational estimates can be judged. The parameter vector is $\vtheta=(\vmu^\top, \vzeta^\top, \vtheta_3^\top)^\top$, where
$\vmu = (\mu_1, \dots, \mu_T)^\top$, $\vzeta = (\zeta_1, \dots, \zeta_T)^\top$, 
$\vtheta_3=(\bar\mu,\tilde{\rho}_\mu, \tilde{\sigma}^2_\mu, \bar\zeta,\tilde{\rho}_\zeta, \tilde{\sigma}^2_\zeta)$ with $\tilde{\rho}_{l} = \log\left( \frac{\rho_{\mu}/0.985}{1 - \rho_{l}/0.985}\right)$, $\tilde{\sigma}^2_{l} = \log(\sigma^2_{l})$ for $l\in\{\mu,\zeta\}$ 
transformed to be unconstrained on the real line.
%\footnote{These transformations are $\tilde{\rho}_{\mu} = \log\left( \frac{\rho_{\mu}/0.985}{1 - \rho_{\mu}/0.985}\right)$, $\tilde{\sigma}^2_{\mu} = \log(\sigma^2_{\mu})$, $\tilde{\rho}_{\zeta} = \log\left( \frac{\rho_{\zeta}/0.985}{1 - \rho_{\zeta}/0.985}\right)$ and $\tilde{\sigma}^2_{\zeta} = \log(\sigma^2_{\zeta})$.} 
A feature of the posterior is that the processes $\{\mu_t\}$ and $\{\zeta_t\}$ exhibit strong  near Markov positive serial dependence. 
Partitioning $\vtheta$ into the three blocks $\vmu$, $\vzeta$ and $\vtheta_3$ allows
 tailoring of the marginal posteriors to capture this.
For both $\vmu$ and $\vzeta$ we adopt marginals M1 with $L_j^{-1}$ a band 1 lower triangular matrix. For example,
when the YJ parameters $\veta=(1,\ldots,1)$, M1 is Gaussian with band one precision matrix $S_j^{-1}L_j^{-\top} L_j^{-1}S_j^{-1}$. For $\vtheta_3$ we also use marginal M1, but with matrix $L_3$ having unconstrained lower triangular elements. 
We use VAs with 3-block GVC-F$p$ and KVC vector copulas to link these three marginals, as well as an independent block posterior (BLK-C) with the same M1 marginals for comparison. Additionally, we use GVC-I to capture the contemporaneous dependence between $\mu_t$ and $\zeta_t$ with an independent block $\vtheta_3$.
The posterior is also approximated by GMF and GC-F5 as benchmark VAs.

\begin{figure}[htbp]
	\centering
	\includegraphics[width=\textwidth]{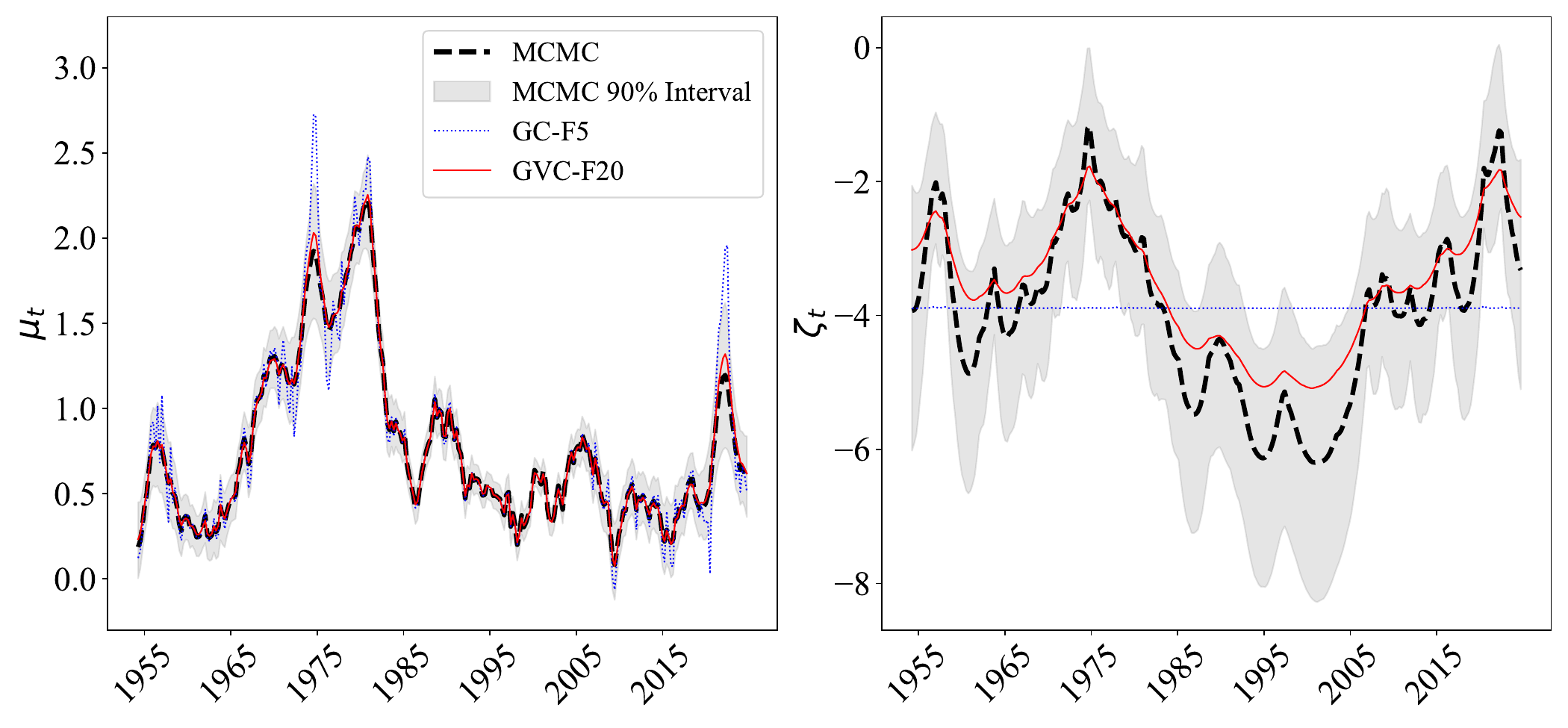}
	\caption{Comparison of the $\vmu$ and $\vzeta$ from the UCSV U.S. inflation example. Posterior means from different methods are plotted.  These are provided for the VAs GC-F5 (blue dotted) and GVC-F20 (red solid), along with the exact posterior computed using MCMC (black dashed). The shaded area represents the 90\% interval derived from the MCMC draws.}
	\label{fig: SVUC_muzeta}
\end{figure}

The size, speed and accuracy of the different VAs are reported in Table~\ref{tab: SVUC_ELBOs}.
Using the marginals M1 with band one $L_j^{-1}$ for $\vmu$ and $\vzeta$ greatly increases the accuracy of
the VA. In contrast, GC-F$p$ (a Gaussian copula with a $p$ factor dependence structure for the entire $\vtheta$ vector) is only slightly more accurate than GMF. For this example, using the vector
copulas to link the blocks improves the accuracy compared to BLK-C, although not as much 
as in the two previous examples. Figure~\ref{fig: SVUC_muzeta} plots the posterior means of $\vmu$ and $\vzeta$ when
computed exactly using MCMC and approximately using GC-F5 and GVC-F20.
GVC-F20 provides an accurate estimate of $\vmu$ and an estimate of $\vzeta$ that falls within
the 90\% posterior probability interval (computed using MCMC). 
In contrast, CG-F5 gives poor estimates of $\vmu$ and $\vzeta$.

%For illustration, the mean and standard deviation of several variational posterior distributions is compared with that estimated by MCMC in Figure~\ref{fig: SVUC_mean} and ~\ref{fig: SVUC_sd}, respectively. It is indicated that GMF cannot approximate the mean and standard deviation of the posterior distribution accurately, notably for $\vzeta$. Additionally, all VI methods in Figure~\ref{fig: SVUC_sd} tend to underestimate the standard deviation, while B2 shows the best performance among them.

%\begin{figure}[htbp]
%	\centering
%	\includegraphics[width=\textwidth]{PLOT/SVUC_mean.eps}
%	\caption{Comparison of the posterior mean from several VAs against MCMC.}
%	\label{fig: SVUC_mean}
%\end{figure}
%
%\begin{figure}[htbp]
%	\centering
%	\includegraphics[width=\textwidth]{PLOT/SVUC_sd.eps}
%	\caption{Comparison of the posterior standard deviation from several VAs against MCMC.}
%	\label{fig: SVUC_sd}
%\end{figure}

\subsection{Example 4: Additive Robust Spline Smoothing}\label{sec:spline}
The fourth application revisits the P-spline regression model in~\citet{ong2018gaussian}, but extended to the additive model 
\begin{equation}\label{eq:pspline}
	y_i=\alpha+\sum_{l=1}^3 g_l(x_{l,i})+e_i\,.
\end{equation}
To robustify the function
estimates to outliers, the errors follow a mixture of two normals with $p(e_i)=0.95\phi(e_i;0,\sigma^2)+0.05\phi(e_i;0,100\sigma^2)$, where $\phi(x;a,b)$ denotes the Gaussian density with mean $a$ and variace $b$. 
%AP: FOR THE SENTENCE ABOVE, I HANGED THIS BECAUSE WE HAVE NOT YET DEFINED LOWER CASE PHI.
Each unknown function is modeled as a P-spline where $g_l(x)=\vb_l(x)^\top \vbeta_l$, with $\vb_l(x)$ a vector of 27 cubic B-spline basis terms evaluated at $x$. To produce smooth
function estimates, the coefficients have proper priors $\vbeta_l | \kappa_l^2, \psi_l \sim \fN(\vzero, \kappa_l^2 P(\psi_l)^{-1})$ where $P(\psi_l)$ is the band one precision matrix of an AR(1) process with autoregressive parameter $\psi_l$. The priors $p(\psi_l)\propto \sone(0<\psi_l<0.99)$, $\alpha \sim \fN(0,100)$, $\sigma^2 \sim \mathcal{IG}(0.01, 0.01)$, and $\kappa_l^2$ is Weibull with shape 0.5 and scale 0.003. For VI the model parameters are transformed to the real line by
setting $\tilde{\sigma}^2 = \log(\sigma^2)$, $\tilde{\kappa}_l^2 = \log(\kappa_l^2)$, and $\tilde{\psi} = \log\left( \frac{\psi_l/0.99}{1 - \psi_l/0.99}\right)$.

A total of $n=1000$ observations are generated from~\eqref{eq:pspline} using the three test functions $g_1(x) = \sin(4\pi x)$, $g_2(x) = 2x - 1$ and $g_3(x) = 0.25 \phi(x; 0.15, 0.05^2) + 0.25\phi(x; 0.6, 0.2^2)$, $\alpha=2$ and $\sigma=0.1$.
The vector $\vx_i=(x_{1,i},x_{2,i},x_{3,i})^\top\in [0,1]^3$ is distributed as a Gaussian copula with parameter matrix
\[
\begin{bmatrix}
	1 & 0.6 & 0.8 \\
	0.6 & 1 & 0.95 \\
	0.8 & 0.95 & 1
\end{bmatrix} \,.
\]

The 11-block partition $\vtheta = (\vbeta_1^\top, \vbeta_2^\top, \vbeta_3^\top, \alpha,
\tilde{\kappa}_1^2, \tilde{\psi}_1, \tilde{\kappa}_2^2, \ldots, \tilde{\psi}_3,
 \tilde{\sigma}^2)^\top$ is used for the three vector copula VAs
GVC-F5, GVC-F20 and KVC-G. Each $\vbeta_j$ has marginal M1 with
$L_j^{-1}$ a band 2 lower triangular matrix (so that $L_j^{-\top} L_j^{-1}$ matches the banded pattern of the prior precision $P_j$) and M1 is used with $d_j=1$ for the singleton blocks. As benchmark  
VAs we use GMF, GC-F5, GC-F20 and 
an independent block posterior (BLK) with the same marginals as the vector copula VAs. 

\begin{figure}[htbp]
	\centering
	\includegraphics[width=\textwidth]{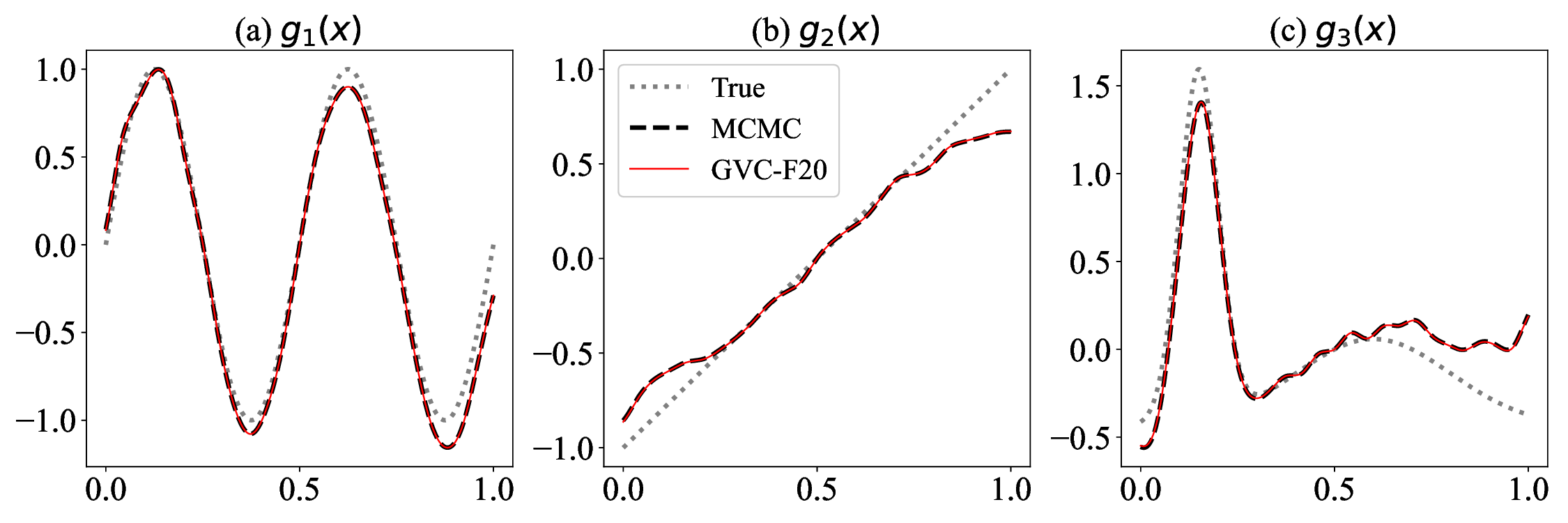}
	\caption{Posterior mean estimates of $g_1$, $g_2$ and $g_3$. The true
		functions (grey dotted line), exact posterior means computed using MCMC (black dashed line) and variational 
		means from GVC-F20 (red solid line) are plotted. The MCMC and GVC-F20 estimates are almost identical and difficult to distinguish visually. Because the functions are not unique up to an intercept in the additive model, we plot the centered functions $g_l(x) - g_l(0.5)$.}
	\label{fig: spline_function}
\end{figure}

The performance of different VAs is reported in Table~\ref{tab:spline}. Among benchmark VAs, GC-F$p$ and BLK are more accurate than GMF. All vector copula VAs improve on BLK, with GVC-F20 being the most accurate VA in this example. Figure~\ref{fig: spline_function} plots the true function, its posterior mean estimates computed exactly using MCMC and approximately using GVC-F20. The latter are very accurate approximations to the exact posterior.

\begin{table}[htbp]
	\begin{center}
	\caption{Size, speed and accuracy of VAs in the spline smoothing example}
	\label{tab:spline}
	\resizebox{1.0\textwidth}{!}{
		\begin{tabular}{lccc}
			\toprule
			\toprule
			& \# of Variational Parameters & Time (s) Per 1000 Steps &  ELBO \\
			\midrule
			\underline{\textit{Benchmark VAs}}: \\
			GMF & 178 & 0.25 & -305.62 \\
			GC-F5 & 702 & 1.20 & -295.22 \\
			GC-F20 & 1857 & 1.22 & -296.28 \\
			BLK-C & 420 & 1.34 & -297.47 \\
			\midrule
			\underline{\textit{Vector Copula Model VAs}}: \\
			GVC-F5  & 856 & 2.92 & -292.47 \\
			GVC-F20  & 2011 & 2.97 & \textbf{-289.07} \\
			KVC-G & 456 & 2.41 & -295.65 \\
			\bottomrule
			\bottomrule
		\end{tabular}
	}
	\end{center}
	Note: The approximations BLK-C, GVC-F$p$ and KVC-G all have the same M1 marginals with band two $L_j^{-1}$ matrices for $\vbeta_1$, $\vbeta_2$ and $\vbeta_3$. The number of variational parameters is $|\vlambda|$. Times are reported for implementation in Python on an Apple M3 Max laptop. %The ELBO values are the median from the last 1000 steps of the SGD algorithm, with higher values indicating greater accuracy.  
	Higher ELBO values indicate greater accuracy, with the largest value in bold. 
\end{table}

\section{Discussion} \label{sec5}
While VI is popular, its effectiveness is largely determined by
the choice of approximation to the posterior distribution. The key is to use an approximation that captures the salient features of a posterior, while also allowing for fast solution of the optimization.
Tractable block posterior approximations are attractive because they allow for tailoring of the marginals. For example,~\cite{ansari18}, \cite{danaher2020} and \cite{bernardi2024} do so when estimating large econometric models, and we show here that tailoring multivariate marginals is important in the UCSV and spline smoothing applications. However, it is difficult to tailor marginals while also capturing inter-block (i.e. between) dependence, so most studies---including those cited above---assume independence between the blocks. Our
study is the first to suggest using vector copula models to also allow for inter-block dependence
and show that this can improve the 
accuracy of the VAs. 

The two learnable transport maps M1 and M2 are used to construct multivariate marginals that are flexible, while remaining parsimonious. They are examples of normalizing flows that are widely used as
VAs in machine learning~\citep{papamakarios2021normalizing}. 
The GVC and KVC vector copula models were proposed by~\cite{fan2023vector} for modeling low dimensional observation data. In this paper we instead use them to construct VAs in higher dimensions. For the GVC, this necessitates careful parameterization of $\Omega$, and we suggest two
approaches (i.e. GVC-F$p$ and GVC-O/I). Both the marginals and vector copulas that we suggest
have convenient generative representations that are suitable for the fast computation of efficient re-parameterized gradients. This is important when implementing SGD algorithms to solve the variational optimization. 

The type of vector copula used should match the likely pattern of between dependence in the posterior. 
For example, when adopting a global-local shrinkage prior, such as the horseshoe in Section~\ref{sec:eg1} or LASSO in Section~\ref{sec:correg}, GVC-I is suitable to capture the dependence between the vectors of continuous model parameters and local hyper-parameters. 
This is a useful observation because global-local shrinkage priors are growing in popularity for the large Bayesian models for which VI is most attractive~\citep{korobilis2022}. 
Other vector copulas, including GVC-F$p$ and KVC-G, are more suitable for capturing the between dependence in other cases. We note that KVC-G exhibits equal dependence between all elements of two blocks, which may be inappropriate for many posteriors. Nevertheless, this is not an issue when capturing the dependence between a singleton block (i.e. a block with a single scalar parameter) and other multivariate blocks. We also stress that VCVI is a highly modular framework. For example, it is possible to use a conventional copula as a multivariate marginal, as with the Gaussian copula marginals formed using M1 or M2 in our applications. It is also possible for one or more of the multivariate marginals to itself be a vector copula; something we discuss in Part~E of the Online Appendix. 

We conclude by noting three promising directions for future work. 
First, because any learnable bijective transport map $T_j$ can be used to define the marginals, an exploration of others that can increase accuracy at low computational cost is worthwhile. This may include the cyclically monotone maps discussed by~\cite{bryan2021multirank} and~\cite{makkuva2020optimal}, or autoregressive flows that have been shown to be effective for posteriors with complex
geometries including multi-modality~\citep{papamakarios2019sequential}.
Second, the GVC can be extended to other elliptical vector copulas using generative representations for elliptical distributions, such as those discussed by~\cite{kano1994}. 
Additionally, the Gaussian nesting copula $C_0$ of the KVC can be replaced by more flexible copulas.
%, such as vine copulas~\citep{aas2009pair}. 
Finally, 
the VCVI method presented here has the potential 
to provide fast and accurate estimates of some challenging large models where mean field VAs can be poor. These include Bayesian neural networks with layer-specific horseshoe priors as in~\cite{JMLR:v20:19-236}, vector autoregressions with hierarchical shrinkage as in~\cite{gefang2023forecasting} and big 
non-nested~\citep{goplerud2022fast} and crossed~\citep{menictas2023streamlined} random effects models.

 %\clearpage
%\newpage
\noindent
\appendix

\section{Kendall Vector Copula}\label{app:kvc}
Let $\vU_j = (U_{j,1}, \ldots, U_{j,d_j})^\top \in [0,1]^{d_j}$ be a random vector. Let $\vU_j$ 
be distributed as an Archimedeam copula $C_j$ with generator function $\psi_j$, so that $C_j(\vU_j) = \psi_j^{-1} ( \sum_{b=1}^{d_j} \psi_j(U_{j,b}))$. \cite{mcneil2009multivariate} give a useful stochastic representation $\vU_j \overset{d}{=} \psi_j^{-1} (R_j \vS_{j})$, where $\vS_j = (S_1, \ldots, S_{d_j})^\top$ is uniformly distributed on the $d_j$-dimensional simplex, and $R_j = \sum_{b=1}^{d_j} \psi_j (U_{j,b})$ has the distribution function
\begin{equation*} \label{FRi}
	F_{R_j}(x)=1-\sum_{b=0}^{d_j-2} \frac{(-1)^b x^b \psi_j^{-1(b)}(x)}{b!}-\frac{(-1)^{d_j-1} x^{d_j-1} \psi_{j+}^{-1(d_j-1)}(x)}{(d_j-1)!}, x \in[0, \infty),
\end{equation*}
with $\psi_j^{-1(b)}$ the $b$-th derivative of the inverse function of $\psi_j$, and the function $\psi_{j+} = \max(0, \psi_j )$.

A hierarchical Kendall copula~\citep{brechmann2014hierarchical} nests the dependence between $\vU_1, \ldots, \vU_M$ in a nesting copula $C_0(V_1, \ldots, V_M)$, where $V_j = K_j(C_j(\vU_j))$, and $K_j$ is the Kendall distribution function of the random variable $C_j(\vU_j)$. \citet{BARBE1996197} give an expression of the Kendall distribution function with the generator $\psi_j$:
\begin{equation*}\label{kdf}
	K_j(t) = t + \sum_{b = 1}^{d_j-1} \frac{(-1)^{b}}{b !} \psi_j(t)^{b} (\psi_j^{-1})^{(b)}(\psi_j(t)).
\end{equation*}

A KVC is a hierarchical Kendall copula with independence cluster copulas $C_j$, for $j = 1, \ldots, M$. An independence copula is an Archimedean copula with generator function $\psi_j(t) = - \ln(t)$,
so that $\vU_j \overset{d}{=} \exp( - R_j \vS_j)$. Moreover, $\psi^{-1}_{j+} (x)= \max(0,e^{-x} ) = e^{-x}$, $\psi_j^{-1(b)}(x) = \psi_{j+}^{-1(b)}(x) = (-1)^{b} e^{-x}$, and
\begin{equation}
	\begin{aligned}
		F_{R_j}(x) &=1-\sum_{b=0}^{d_j-1} \frac{(-1)^b x^b \psi_j^{-1(b)}(x)}{b!} \\
		&= 1-\sum_{b=0}^{d_j-1} \frac{(-1)^b x^b (-1)^{b} e^{-x}}{b!} \\
		&= 1-\sum_{b=0}^{d_j-1} \frac{x^b e^{-x}}{b!}, x \in[0, \infty),
	\end{aligned}
\end{equation}
which is the distribution function of the Erlang distribution with the scale parameter 1 and shape parameter $d_j$. The Kendall distribution function of an independence copula is tractable:
\begin{eqnarray}
		K_j(t) &= t + \sum_{b = 1}^{d_j-1} \frac{(-1)^{b}}{b !} (-\ln t)^{b} (-1)^{b} \exp(- (-\ln t)) \nonumber \\
		&= t + \sum_{b = 1}^{d_j-1} \frac{t(-\ln t)^{b}}{b !}  \nonumber \\
		&= \sum_{b = 0}^{d_j-1} \frac{t(-\ln t)^{b}}{b !}\,.\label{kdfic}
\end{eqnarray}
For a KVC, the random variable
$V_j = K_j(C_j(\vU_j)) = K_j (\psi_j^{-1} ( \sum_{b=1}^{d_j} \psi_j(U_{j,b}))) = K_j (\psi_j^{-1}(R_j)) = K_j(\exp(-R_j))$. Using the expression at~\eqref{kdfic}, we have $V_j =  \sum_{b = 0}^{d_j-1} \frac{R_j^{b} \exp(-R_j) }{b !} = 1 - F_{R_j}(R_j)$.

These derivations provide a way to sample from the KVC. First sample $\vv = (v_1, \ldots, v_M)^\top~\sim C_0$, and then for $j=1,\ldots,M$ set $r_j = F_{R_j}^{-1}(1 - v_j)$, where $F_{R_j}^{-1}$ is the quantile function of the Erlang distribution as specified above. Then,
for $j=1,\ldots,M$, sample the vector of independent exponentials
$\ve=(e_1, \ldots, e_{d_j})^\top$ and compute $\vs_j = (s_1, \ldots, s_{d_j})^\top$ with $\vs_j = \frac{\ve_j}{||\ve_j||}$. Finally, set $\vu_j = \exp(- r_j \vs_j)$, and $\vu = (\vu_1^\top, \ldots, \vu_M^\top)^\top$ is a draw from
a KVC; see Algorithm~\ref{alg: sample from kvc}.

\clearpage
\singlespacing
\bibliography{core/references}
 
\clearpage
\onehalfspacing
\setcounter{page}{1}
\begin{center}
	{\bf \Large{Online Appendix for ``Vector Copula Variational Inference and Dependent Block Posterior Approximations''}}
\end{center}

\vspace{10pt}

\setcounter{figure}{0}
\setcounter{table}{0}
\setcounter{section}{0}
\setcounter{equation}{0}
\setcounter{algorithm}{1}
\renewcommand{\thetable}{A\arabic{table}}
\renewcommand{\thefigure}{A\arabic{figure}}
\renewcommand{\thealgorithm}{\Alph{section}\arabic{algorithm}}
\renewcommand{\thesection}{Part~\Alph{section}}
\renewcommand{\theequation}{A\arabic{equation}}

% ------------------------------  contents ---------------------------------------------------
\noindent This Online Appendix has five parts:

\begin{itemize}
	\item[] {\bf Part~A}: Notational conventions and matrix differentiation rules used
		\item[] {\bf Part~B}: Details of GVC-O
	\begin{itemize}
		\item[] B.1: Inverse of $\Omega$
		\item[] B.2: Analytical gradients of A3 (GVC-I \& M1)
	\end{itemize}
	\item[] {\bf Part~C}: Efficient re-parameterized gradients for VCVI and KVC-G
	\begin{itemize}
		\item[] C.1: VCVI
		\item[] C.2: KVC-G
	\end{itemize}
	\item[] {\bf Part~D}: Additional empirical information
	\item[] {\bf Part~E}: Nested vector copula density
	\begin{itemize}
		\item[] E.1: Re-parameterization trick
		\item[] E.2: Empirical results
	\end{itemize}
	\end{itemize}
\clearpage

% ------------------------------------------------------------------------------------------
\noindent {\bf \large{Part~A: Notational conventions and matrix differentiation rules used}}\\

\noindent We outline the notational conventions that we adopt in computing
derivatives throughout the paper. For a $d$-dimensional vector valued function $g(\bm x)$ of an $n$-dimensional
argument $\bm x$, $\frac{\partial g}{\partial \bm x}$ is the $d\times n$ matrix with element $(i,j)$ $\frac{\partial g_i}{\partial x_j}$.  This means for a scalar $g(\bm x)$, $\frac{\partial g}{\partial \bm x}$ is
a row vector.  When discussing the gradients we also sometimes write $\nabla_x g(\bm x)=\left[ \frac{\partial g}{\partial \bm x} \right]^\top$, which is a column vector.
When the function $g(\bm x)$ or the argument $\bm x$ are matrix valued, then $\frac{\partial g}{\partial \bm x}$ is taken to 
mean $\frac{\partial \text{vec}(g(\bm x))}{\partial \text{vec}(\bm x)}$, where $\text{vec}(A)$ denotes the vectorization of a matrix $A$ obtained by stacking its columns one
underneath another.

When an arithmetic operator is applied to a vector, such as subtraction, exponential and the square root, it is applied to each element in that vector.
\clearpage

% ------------------------------------------------------------------------------------------

\noindent {\bf \large{Part~B: Details of GVC-O}}\\

\noindent \textbf{B.1: Inverse of $\Omega$}

\noindent We use the block matrix inverse formula:
\begin{equation*}
	\begin{bmatrix}
		A & B \\
		C & D
	\end{bmatrix}^{-1} = 
	\begin{bmatrix}
		A^{-1} + A^{-1} B (D - C A^{-1} B)^{-1} C A^{-1} & -A^{-1} B (D - C A^{-1} B)^{-1} \\
		-(D - C A^{-1} B)^{-1} C A^{-1} & (D - C A^{-1} B)^{-1}
	\end{bmatrix}
\end{equation*}
for the matrix
\begin{equation*}%\label{eqapp: Omega2}
	\Omega = \begin{bmatrix}
		I_{d_1} & Q_1 \Lambda Q_2^\top \\
		Q_2 \Lambda Q_1^\top & I_{d_2}
	\end{bmatrix}\,.
\end{equation*}

Under the assumption that $d_1 \leq d_2$, we have shown that $(D - C A^{-1} B)^{-1} = (\Omega_2/I_{d_1})^{-1} = Q_2\left( I_{\tilde{d}} - \Lambda^2 \right)^{-1} Q_2^\top$. As a result, we have 
\begin{equation*}
	\begin{aligned}
		& A^{-1} + A^{-1} B (D - C A^{-1} B)^{-1} C A^{-1} \\ 
		=& I_{d_1} + Q_1 \Lambda Q_2^\top  Q_2 \left( I_{\tilde{d}} - \Lambda^2 \right)^{-1} Q_2^\top Q_2 \Lambda Q_1^\top \\
		=& I_{d_1} + Q_1 \Lambda \left( I_{\tilde{d}} - \Lambda^2 \right)^{-1} \Lambda Q_1^\top
	\end{aligned}
\end{equation*}
and
\begin{equation*}
	\begin{aligned}
		& -A^{-1} B (D - C A^{-1} B)^{-1} \\ 
		=& -Q_1 \Lambda Q_2^\top Q_2\left( I_{\tilde{d}} - \Lambda^2 \right)^{-1} Q_2^\top \\
		=& -Q_1 \Lambda \left( I_{\tilde{d}} - \Lambda^2 \right)^{-1} Q_2^\top \,.
	\end{aligned}
\end{equation*}

%\begin{equation*}
%	\begin{aligned}
	%	& -(D - C A^{-1} B)^{-1} C A^{-1} \\ 
	%	&=  -  Q_2\left( I_{\tilde{d}} - \Lambda^2 \right)^{-1} Q_2^\top Q_2 \Lambda Q_1^\top \\
	%	=& -  Q_2\left( I_{\tilde{d}} - \Lambda^2 \right)^{-1} \Lambda Q_1^\top,
	%	\end{aligned}
%\end{equation*}
%and
%\begin{equation*}
%	\begin{aligned}
	%		& (D - C A^{-1} B)^{-1} \\ 
	%		=& Q_2\left( I_{\tilde{d}} - \Lambda^2 \right)^{-1} Q_2^\top.
	%	\end{aligned}
%\end{equation*}

\noindent Thus, 
\begin{equation*}%\label{eqapp: Omega2inv}
	\Omega^{-1} =  \begin{bmatrix}
		I_{d_1} + Q_1 \Lambda (I_{\tilde{d}} - \Lambda^2)^{-1} \Lambda Q_1^\top & -Q_1 \Lambda (I_{\tilde{d}} - \Lambda^2)^{-1} Q_2^\top \\
		- Q_2 (I_{\tilde{d}} - \Lambda^2)^{-1}\Lambda Q_1^\top& Q_2 (I_{\tilde{d}} - \Lambda^2)^{-1} Q_2^\top
	\end{bmatrix} \,.
\end{equation*}

When $d_1 = d_2 = \tilde{d}$ and $Q_1 = Q_2 = I_{\tilde{d}}$,
\begin{equation*}%\label{eqapp: Omega2t}
	\Omega = \begin{bmatrix}
		I_{\tilde{d}} & \Lambda \\
		\Lambda & I_{\tilde{d}}
	\end{bmatrix}
\end{equation*}
and
\begin{equation*}%\label{eqapp: Omega2invt}
	\Omega^{-1} = \begin{bmatrix}
		I_{d_1} + \Lambda (I_{\tilde{d}} - \Lambda^2)^{-1} \Lambda & \Lambda (I_{\tilde{d}} - \Lambda^2)^{-1} \\
		- (I_{\tilde{d}} - \Lambda^2)^{-1}\Lambda & (I_{\tilde{d}} - \Lambda^2)^{-1}
	\end{bmatrix} \,.
\end{equation*}

\noindent \textbf{B.2: Analytical gradients of A3 (GVC-I \& M1)}

\noindent To implement this method, we first sample $\vz = (\vz_1^\top, \vz_2^\top, z_3)^\top$ from $\fN(\vzero_{m+m+1}, \widetilde{\Omega})$. Here, 
\begin{equation*}
	\widetilde{\Omega} = \begin{bmatrix}
		I_{m} & \Lambda & \vzero_{m \times 1} \\
		\Lambda & I_{m} & \vzero_{m \times 1} \\
		\vzero_{1 \times m} & \vzero_{1 \times m} & 1
	\end{bmatrix}\,,
\end{equation*}
and $\vz$ can be sampled by the re-parameterization trick
\begin{equation*}
	\vz = \begin{bmatrix}
		I_m & \vzero_{m \times m} & \vzero_{m \times 1} \\
		\Lambda & \sqrt{I_m - \Lambda^2} & \vzero_{m \times 1} \\
		\vzero_{1 \times m} & \vzero_{1 \times m} & 1
	\end{bmatrix} \vepsilon = 
	\begin{bmatrix}
		\vepsilon_1 \\
		\vl \circ \vepsilon_1 + \sqrt{\bm{1}_m - \vl^2 } \circ \vepsilon_2 \\
		\epsilon_3
	\end{bmatrix} \,,
\end{equation*}
where $\vepsilon = (\vepsilon_1^\top, \vepsilon_2^\top, \epsilon_3)^\top \sim \fN(0, I_{m+m+1})$ and the diagonal elements of $\Lambda$ is represented by the vector $\vl$ which has constrained values from -1 to 1. 

Then $(\vu_1^\top, \vu_2^\top)^\top = \Phi\left( (\vz_1^\top, \vz_2^\top)^\top ; I_{2m}\right)$ follows a GVC-I, and $u_3 = \Phi\left( z_3 ; I_{1}\right)$ is independent of $(\vu_1^\top, \vu_2^\top)^\top$. The multivariate margins are constructed by $\vtheta_i = \vb_i + S_i \Phi_i^{-1}(\vu_i) = \vb_i + S_i \vz_i$ for $i = 1,2,3$. Thus, we note that $\vtheta$ can be simply represented by $\vtheta = \vb + \vs \circ \vz$, where $\vb = (\vb_1^\top, \vb_2^\top, b_3)^\top$, $\vs$ is the leading diagonal of the diagonal matrix $S$ and $\circ$ is the Hadamard product. 

The vectors $\vs$ and $\vl$ have constrained values. We use $\vs = \tilde{\vs}^2$ and $\vl = \frac{2}{\exp(-\tilde{\vl}) + 1} - 1$ such that $\tilde{\vs}$ and $\tilde{\vl}$ are two vectors defined on the real line. As the result, the variational parameters are $\vlambda = (\tilde{\vl}^\top, \tilde{\vs}^\top, \vb^\top)^\top$. 

We are interested in the re-parameterized gradients:
\begin{equation*}%\label{eqapp: elbogradient}
	\nabla_{\vlambda} \fL (\vlambda) = \Exp_{f_\epsilon} \left\{ \left[ \frac{d\vtheta}{d \vlambda} \right]^\top \nabla_{\vtheta} \left(\log h(\vtheta) -\log q(\vtheta)\right)\right\} .
\end{equation*}
Here, we evaluate $\nabla_{\vtheta} \log h(\vtheta)$ by automatic differentiation, and $\nabla_{\vtheta}\log q(\vtheta) = - S^{-1}\cdot \widetilde{\Omega}^{-1} \cdot  \vtheta$.
It is straightforward to see that 
\begin{equation*}
	\begin{aligned}
		\frac{d\vtheta}{d\vb} &= I_{2m+1} \,, \\
		\frac{d\vtheta}{d\tilde{\vs}} &= Diag(2 \tilde{\vs} \circ \vz) \,, \\
		\frac{d\vtheta}{d\tilde{\vl}} &= \frac{d\vtheta}{d\vz} \frac{d\vz}{d\vl} \frac{d \vl}{d \tilde{\vl}} \,.
	\end{aligned}
\end{equation*}
For the last term, we have
\begin{equation*}
	\frac{d\vz}{d\vl} = 
	\begin{bmatrix}
		\vzero_{m \times m} \\
		\frac{d\vz_2}{d\vl} \\
		\vzero_{1 \times m}
	\end{bmatrix} \,,
\end{equation*}
where
\begin{equation*}
	\frac{d\vz_2}{d\vl} = Diag \left( \vepsilon_1  - \frac{\vl}{\sqrt{\bm{1}_m - \vl^2}} \circ \vepsilon_2 \right) \,.
\end{equation*}
Moreover, 
\begin{equation*}
	\frac{d\vtheta}{d\vz} = Diag (\tilde{\vs}^2)
\end{equation*}
and
\begin{equation*}
	\frac{d\vl}{d\tilde{\vl}} = Diag \left\{ \frac{2\exp(-\tilde{\vl})}{(\exp(-\tilde{\vl}) + 1)^2} \right\} \,.
\end{equation*}
As a result,
\begin{equation*}
	\frac{d\vtheta}{d\tilde{\vl}} = Diag (\tilde{\vs}^2) \cdot 
	\begin{bmatrix}
		\vzero_{m \times m} \\
		Diag \left( \vepsilon_1  - \frac{\vl}{\sqrt{\bm{1}_m - \vl^2}} \circ \vepsilon_2 \right) \\
		\vzero_{1 \times m}
	\end{bmatrix} \cdot
	Diag \left\{ \frac{2\exp(-\tilde{\vl})}{(\exp(-\tilde{\vl}) + 1)^2} \right\} \,.
\end{equation*}

Thus,
\begin{equation*}
	\begin{aligned}
		\nabla_{\vb}\fL(\vlambda) &= \nabla_{\vtheta} \left(\log h(\vtheta) -\log q(\vtheta)\right)\,, \\
		\nabla_{\tilde{\vs}}\fL(\vlambda) &= 2 \tilde{\vs} \circ \vz \circ \nabla_{\vtheta} \left(\log h(\vtheta) -\log q(\vtheta)\right)\,, \\
	\end{aligned}
\end{equation*}
and
\begin{equation}\label{eq: dELBOdtildel}
	\begin{aligned}
		\nabla_{\tilde{\vl}}\fL(\vlambda) =& \left\{ Diag (\tilde{\vs}^2) \cdot 
		\begin{bmatrix}
			\vzero_{m \times m} \\
			Diag \left( \vepsilon_1  - \frac{\vl}{\sqrt{\bm{1}_m - \vl^2}} \circ \vepsilon_2 \right) \\
			\vzero_{1 \times m}
		\end{bmatrix} \cdot
		Diag \left\{ \frac{2\exp(-\tilde{\vl})}{(\exp(-\tilde{\vl}) + 1)^2} \right\}
		\right\}^\top \\
		& \cdot \nabla_{\vtheta} \left(\log h(\vtheta) -\log q(\vtheta)\right) \\
		=& Diag \left\{ \frac{2\exp(-\tilde{\vl})}{(\exp(-\tilde{\vl}) + 1)^2} \right\} \cdot 
		\begin{bmatrix}
			\vzero_{m \times m} & Diag \left( \vepsilon_1  - \frac{\vl}{\sqrt{\bm{1}_m - \vl^2}} \circ \vepsilon_2 \right) & \vzero_{m \times 1}
		\end{bmatrix} \\
		& \cdot Diag (\tilde{\vs}^2) \cdot 
		\nabla_{\vtheta} \left(\log h(\vtheta) -\log q(\vtheta)\right) \\
		=& \frac{2\exp(-\tilde{\vl})}{(\exp(-\tilde{\vl}) + 1)^2}  \circ \left( \vepsilon_1  - \frac{\vl}{\sqrt{\bm{1}_m - \vl^2}} \circ \vepsilon_2 \right) \circ \tilde{\vs}^2_{m:2m} \circ \left[  \nabla_{\vtheta} \left(\log h(\vtheta) -\log q(\vtheta)\right) \right]_{m:2m} \,,
	\end{aligned}
\end{equation}
where $m:2m$ denotes keeping the elements from element $m$ to $2m$ in a vector. The last equality in~(\ref{eq: dELBOdtildel}) can be seen by treating matrix multiplications as row operations or column operations.  

To conclude, we underline that only Hadamard products between vectors are required to update the variational parameters in A3 (except for $\nabla_{\vtheta}logh(\vtheta)$).

\clearpage

% ---------------------------------------------------------------
\noindent {\bf \large{Part~C: Efficient re-parameterized gradients for VCVI and KVC-G}}\\

\noindent \textbf{C.1: VCVI}

\noindent When we use a vector copula density as the variational density, equation~\eqref{eq: elboq} can be written as
\begin{equation}\label{eq: elboqvc}
	\fL(q) = \Exp_{q}\left[\log h(\vtheta)-\log c_v(\vu;\vlambda_{\mbox{\footnotesize vc}}) - \sum_{j=1}^{M} \log q_j(\vtheta_j;\vlambda_j)\right] \,,
\end{equation}
where $c_v$ and $q_j$ are the probability density functions of $\vu$ and $\vtheta_j$, respectively. 
%Here, $\vlambda_{\mbox{\footnotesize vc}}$ does not contribute to $logq_j(\vtheta_j;\vlambda_j)$ because $\vtheta_j = T_j(\vu_j; \vlambda_j)$ and the variables within $\vu_j$ are distributed as independent uniform distributions. Additionally, we note that $\vu = g_1\left(\vepsilon, \vlambda_{\mbox{\footnotesize vc}}\right)$ will always be generated before $\vtheta$ in VCVI, and thus $\vlambda_{\mbox{\footnotesize marg}}$ does not contribute to $\log c_v(\vu;\vlambda_{\mbox{\footnotesize vc}})$. 
Throughout the paper, $\Exp_{f}[g]$ denotes the expectation of a function $g$ with respect to 
random vector $\bm{X}\sim f$ with $f$ the density function of $\bm{X}$.
Equation~\eqref{eq: elboqvc} can be simplified as
\begin{equation}\label{eq: elboqvc2}
	\fL(q) = \Exp_{q}\left[\log h(\vtheta) \right] - \Exp_{c_v} \left[ \log c_v(\vu;\vlambda_{\mbox{\footnotesize vc}}) \right] - \sum_{j=1}^{M} \Exp_{q_j} \left[\log q_j(\vtheta_j;\vlambda_j)\right] \,.
\end{equation}
Such a simplification from~\eqref{eq: elboqvc2} enables us to evaluate the gradient efficiently. Specifically, we have
\begin{equation*}
	\begin{aligned}
	\nabla_{\vlambda_j} \Exp_{q_j} \left[logq_j(\vtheta_j;\vlambda_j)\right] &= \Exp_{q_j} \left[ \left(\frac{d\vtheta_j}{d\vlambda_j} \right)^\top \nabla_{\vtheta_j} \log q_j(\vtheta_j;\vlambda_j) + \nabla_{\vlambda_j}logq_j(\vtheta_j;\vlambda_j)\right] \\ &=  \Exp_{q_j} \left[ \left(\frac{d\vtheta_j}{d\vlambda_j} \right)^\top \nabla_{\vtheta_j} \log q_j(\vtheta_j;\vlambda_j)\right]
	\end{aligned}
\end{equation*}
and
\begin{equation*}
	\begin{aligned}
	\nabla_{\vlambda_{\mbox{\footnotesize vc}}} \Exp_{c_v} \left[ \log c_v(\vu;\vlambda_{\mbox{\footnotesize vc}}) \right] &=  \Exp_{c_v} \left[ \left(\frac{d\vu}{d\vlambda_{\mbox{\footnotesize vc}}} \right)^\top \nabla_{\vu} \log c_v(\vu;\vlambda_{\mbox{\footnotesize vc}}) + \nabla_{\vlambda_{\mbox{\footnotesize vc}}} \log c_v(\vu;\vlambda_{\mbox{\footnotesize vc}}) \right] \\ &= \Exp_{c_v} \left[ \left(\frac{d\vu}{d\vlambda_{\mbox{\footnotesize vc}}} \right)^\top \nabla_{\vu} \log c_v(\vu;\vlambda_{\mbox{\footnotesize vc}})\right]
	\end{aligned}
\end{equation*}
because of the log derivative trick.

Thus, the re-parameterized gradients of~\eqref{eq: elboqvc2} are
\begin{equation*}
\nabla_{\vlambda_{\mbox{\footnotesize vc}}} \fL (\vlambda) = \Exp_{f_\epsilon} \left\{ \left[ \frac{d\vtheta}{d \vlambda_{\mbox{\footnotesize vc}}} \right]^\top \nabla_{\vtheta} \left( \log h(\vtheta) - \sum_{j=1}^{M}logq_j(\vtheta_j;\vlambda_j) \right) - \left[\frac{d\vu}{d\vlambda_{\mbox{\footnotesize vc}}} \right]^\top \nabla_{\vu} \log c_v(\vu;\vlambda_{\mbox{\footnotesize vc}}) \right\} \,,
\end{equation*}

\begin{equation} \label{eq: dmarg}
\nabla_{\vlambda_{\mbox{\footnotesize marg}}} \fL (\vlambda) = \Exp_{f_\epsilon} \left\{\left[ \frac{d\vtheta}{d \vlambda_{\mbox{\footnotesize marg}}} \right]^\top \nabla_{\vtheta} \left( \log h(\vtheta) - \sum_{j=1}^{M}logq_j(\vtheta_j;\vlambda_j) \right)\right\}\,,
\end{equation}
where $\vtheta = g_2\left(g_1\left(\vepsilon, \vlambda_{\mbox{\footnotesize vc}}\right), \vlambda_{\mbox{\footnotesize marg}}\right)$ and $\vu = g_1\left(\vepsilon, \vlambda_{\mbox{\footnotesize vc}}\right)$.
~\\

\noindent \textbf{C.2: KVC-G}

\noindent We can further simplify $\Exp_{f_\epsilon}\left[ c_v(\vu;\vlambda_{\mbox{\footnotesize vc}})\right]$ in the context of KVC-G, where we have $f_{\epsilon} = p(\vepsilon_1, \ve) = p(\vepsilon_1)p(\ve)$ with $\ve = (\ve_1^\top, \dots, \ve_M^\top)^\top$. The density of a Gaussian copula is
\begin{equation*}
	c_0^{Ga}(v_1, \cdots, v_M; \bar{\Omega}) = |\overline{\Omega}|^{-1 / 2} \exp \left\{-\frac{1}{2} \vkappa^{\top}\left(\overline{\Omega}^{-1}-I_M\right) \vkappa\right\},
\end{equation*}
where $\vkappa = \underline{\Phi}^{-1}(\vv)$ and $\vv = (v_1, \dots, v_M)^\top$. Let $\tilde{G}$ be the lower triangular Cholesky factor of $\bar{\Omega}$. To sample from the Gaussian copula, we set $\vv = \underline{\Phi}(\tilde{G}\vepsilon_1)$ with $\vepsilon_1 \sim \fN(\vzero, I_M)$.
The log density under the expectation gives:
\begin{equation*}
	\begin{aligned}
		&  E_{p(\vepsilon_1,\ve)}\left[\log  c_v(\vu_1, \ldots, \vu_M)\right] \\
		=&  E_{p(\vepsilon_1,\ve)}\left[\log  c_0^{Ga}(v_1, \ldots, v_M; \bar{\Omega})\right] \\ =& \int_{\ve} \int_{\vepsilon_1}\log  c_0^{Ga}(v_1, \ldots, v_M; \bar{\Omega}) p(\vepsilon_1) p(\ve) d \vepsilon_1 d \ve \\
		=& \int_{\vepsilon_1}\log  c_0^{Ga}(v_1, \ldots, v_M; \bar{\Omega}) p(\vepsilon_1) d \vepsilon_1   \int_{\ve} p(\ve) d\ve \\
		=& E_{p(\vepsilon_1)} \left[\log  c_0^{Ga}(v_1, \ldots, v_M; \bar{\Omega})\right]  \\
		=& E_{p(\vepsilon_1)} \left[\log  \left(|\overline{\Omega}|^{-1 / 2} \exp \left\{-\frac{1}{2} (\tilde{G}\vepsilon_1)^{\top}\left(\overline{\Omega}^{-1}-I_M\right) \tilde{G}\vepsilon_1 \right\} \right)\right] \\
		=& E_{p(\vepsilon_1)} \left[ -\frac{1}{2}log|\tilde{G} \tilde{G}^\top| -\frac{1}{2}(\tilde{G} \vepsilon_1)^\top (\tilde{G} \tilde{G}^\top)^{-1}\tilde{G}\vepsilon_1 + \frac{1}{2}(\tilde{G} \vepsilon_1)^\top \tilde{G}\vepsilon_1\right] \\
		=& E_{p(\vepsilon_1)} \left[ -\frac{1}{2}log|\tilde{G} \tilde{G}^\top| -\frac{1}{2}\vepsilon_1^\top \vepsilon_1 + \frac{1}{2}\vepsilon_1^\top \tilde{G}^\top \tilde{G}\vepsilon_1\right] \\
		=& -\frac{1}{2}E_{p(\vepsilon_1)} \left[\log |\tilde{G}|^2 \right] - \frac{1}{2} E_{p(\vepsilon_1)} \left[ \vepsilon_1^\top \vepsilon_1 \right] + \frac{1}{2} E_{p(\vepsilon_1)} \left[ \vepsilon_1^\top \tilde{G}^\top \tilde{G} \vepsilon_1 \right] \\
		=& -\log |\tilde{G}| - \frac{1}{2} \cdot M + \frac{1}{2} trace(\overline{\Omega}^\top)\\
		=& -\log |\tilde{G}| - \frac{1}{2} \cdot M + \frac{1}{2} \cdot M \\
		=& -\log |\tilde{G}| \,,
	\end{aligned}
\end{equation*}
where $E_{p(\vepsilon_1)} \left[ \vepsilon_1^\top \tilde{G}^\top \tilde{G} \vepsilon_1\right]  =  trace(\overline{\Omega}^\top)$ is a result based on equation (318) from the {\em Matrix Cookbook} of Petersen and Pedersen~(2008)
%~\citep{petersen2008matrix}.

 Thus, the ELBO function for KVC can be expressed as:
\begin{equation*}
	\fL(q) = \Exp_{q} \left[\log h(\vtheta) \right] - \sum_{i=1}^{M} \Exp_{q_j} \left[\log  q_j(\vtheta_j;\vlambda_j) \right] +\log |\tilde{G}|\,.
\end{equation*}
Such a result simplifies $\nabla_{\vlambda_{\mbox{\footnotesize vc}}} \fL (\vlambda)$ for KVC:
\begin{equation}\label{eq: dvc}
	\nabla_{\vlambda_{\mbox{\footnotesize vc}}} \fL (\vlambda) = \Exp_{f_\epsilon} \left\{ \left[ \frac{d\vtheta}{d \vlambda_{\mbox{\footnotesize vc}}} \right]^\top \nabla_{\vtheta} \left( \log h(\vtheta) - \sum_{j=1}^{M}\log q_j(\vtheta_j;\vlambda_j) \right) \right\} + \nabla_{\vlambda_{\mbox{\footnotesize vc}}} \log  | \tilde{G} |\,.
\end{equation}

To conclude, we use~\eqref{eq: dmarg} and~\eqref{eq: dvc} as the re-parameterized gradients for KVC-G.
\clearpage

	\begin{sidewaystable}[htbp]
			\noindent {\bf \large{Part~D: Additional empirical information}} \\
	\begin{center}
		\caption{Computational time per 1000 steps (in proportion to GMF) for different VAs in the logistic regression}
		\begin{tabular}{lccccccccc}
			\toprule
			\toprule
			& \multicolumn{4}{c}{Low-Dimensional Real} & \multicolumn{2}{c}{High-Dimensional Real}  & \multicolumn{3}{c}{High-Dimensional Simulated} \\
			\cmidrule(lr){2-5} \cmidrule(lr){6-7} \cmidrule(lr){8-10}
			Dataset Name & krkp & mushroom & spam & iono & cancer & qsar & simu1 & simu2 & simu3 \\
			Dataset Size ($n$x$m$)& 3196x38 & 8124x96 & 4601x105 & 351x112 & 42x2001 & 8992x1024 & 5Kx10K & 20Kx10K & 100Kx10K \\
			\midrule
			\multicolumn{10}{l}{\underline{\textit{Benchmark VAs}}:} \\
			GMF & 0.40 & 0.50 & 0.47 & 0.34 & 0.52 & 2.49 & 12.60 & 53.64 & 285.19 \\
			\cdashline{2-10} \\
			GMF (AG) & 0.85 & 0.90 & 0.90 & 0.81 & 0.88 & 0.92 & 0.98 & 0.99 & 0.99 \\
			G-F5 & 1.97 & 1.97 & 2.08 & 2.35 & 40.53 & 4.22 & 56.81 & 14.41 & 3.32 \\
			GC-F5 & 3.51 & 3.27 & 3.52 & 4.38 & 42.48 & 4.61 & 55.01 & 14.53 & 3.52 \\
			G-20 & 2.00 & 2.03 & 2.25 & 2.69 & 40.48 & 3.56 & 58.05 & 14.51 & 3.53 \\
			GC-20 & 3.52 & 3.39 & 3.65 & 4.76 & 43.41 & 4.03 & 59.43 & 14.76 & 3.52 \\
			BLK & 3.19 & 2.89 & 3.07 & 3.83 & 90.65 & 8.72 & 94.80 & 23.25 & 5.69 \\
			BLK-C & 4.49 & 4.26 & 4.53 & 5.82 & 93.60 & 9.48 & 94.63 & 23.47 & 5.70 \\
			\midrule
			\multicolumn{10}{l}{\underline{\textit{Vector Copula Model VAs}}:} \\
			A1: (GVC-F5 \& M1) & 4.15 & 5.28 & 6.13 & 8.35 & 1494.00 & 69.51 & CI & CI & CI \\
			A2: (GVC-F5 \& M1+YJ) & 5.72 & 6.73 & 7.82 & 10.33 & 1500.01 & 70.11 & CI & CI & CI \\
			\cdashline{2-10} \\
			A3: (GVC-I \& M1) & 1.41 & 1.35 & 1.39 & 1.60 & 1.44 & 1.09 & 1.09 & 1.02 & 0.99 \\
			A3 (AG): (GVC-I \& M1) & 1.14 & 1.12 & 1.14 & 1.24 & 1.17 & 0.90 & 1.02 & 1.00 & 1.00 \\
			A4: (GVC-I \& M1+YJ) & 2.70 & 2.47 & 2.57 & 3.28 & 4.10 & 1.44 & 1.53 & 1.12 & 1.01 \\
			\cdashline{2-10} \\
			A5:  (GVC-I \& M2) & 3.39 & 3.22 & 3.50 & 4.44 & 96.75 & 9.52 & 102.14 & 26.72 & 6.29 \\
			A6: (GVC-I \& M2+YJ) & 4.94 & 4.55 & 4.84 & 6.45 & 100.04 & 9.97 & 103.24 & 26.79 & 6.28 \\
%			\cdashline{2-10} \\
%			A7 & 4.18 & 3.74 & 3.83 & 4.99 & 5.54 & 1.75 & 1.66 & 1.17 & 1.01 \\
			\bottomrule
			\bottomrule
		\end{tabular}
	\end{center}
	Note: The computational time (seconds) for GMF is reported in the table. The time of other VAs is divided by that from GMF. All VAs are trained with double precision on the CPU of Apple M3 Max. The gradients to update variational parameters are evaluated by 1000 times (100 times for simulated datasets, and then scale to 1000). The gradients are computed by Pytorch automatically except for GMF (AG) and A3 (AG). ``AG'' represents analytical gradients and ``CI'' represents computationally infeasible.
\end{sidewaystable}

% -------------------------------------------------------------------------------------------------
\begin{sidewaystable}[htbp]
	\begin{center}
		\caption{List of Example Applications and Dimension of Posterior}
		\label{tab: list_ex}	
		\begin{tabular}{llll}
			\toprule
			Example & Model Type & Prior Type & Posterior Dimension $d$ \\
			\toprule
			E.g.1 (low-dim datasets) & Sparse Logistic Regression & Horseshoe & \\
			\quad  {\em krkp, mushroom, spam, iono} & & & 77, 193, 211, 225 \\
			E.g.1 (High-dim datasets) & Sparse Logistic Regression & Horseshoe & \\
			\quad  {\em cancer, qsar} & & & 4003, 2049 \\
			E.g.1 (Simulated datasets) & Sparse Logistic Regression & Horseshoe & \\
			\quad  {\em simu1, simu2, simu 3} & & & 20001, 20001, 20001 \\
			E.g. 2 & Regularized Correlation Matrix & Bayesian LASSO & \\
			\quad  $r=5$, 10, 20, 30, 49 & & & 21, 91, 381, 871, 2353 \\
			E.g. 3 & UCSV for U.S. Inflation & Autoregressive & 572 \\
			E.g. 4 & Additive Robust P-Spline & Autoregressive & 89 \\
			\bottomrule
		\end{tabular}
	\end{center}
\end{sidewaystable}
\clearpage

% -----------------------------------------------------------------------------------------------------------------
\noindent {\bf \large{Part~E: Nested vector copula density}}\\

\noindent \textbf{E.1: Re-parameterization trick}

\noindent When one or more multivariate marginals of a vector copula model are themselves vector copula models, then this defines a ``nested'' vector copula. For example, consider the case where $M=3$ and the nested vector copula is constructed from two bivariate vector copulas labeled ``a'' and ``b''. The 
first has density $c_v^a$ with marginals of size $d_1^a=d_1+d_2,\, d_2^a=d_3$, and the 
second has density $c_v^b$ with marginals of size $d_1^b=d_1,\, d_2^b=d_2$. Then, the VA density when $C_v^a$ nests $C_v^b$ is
\begin{eqnarray*}
q_\lambda(\vtheta)&=&c_v^a(\tilde{\vu}_{12},\tilde{\vu}_3;\vlambda_{\mbox{\footnotesize vc}}^a)q_{12}(\vtheta_1,\vtheta_2)q_3(\vtheta_3)\\
&=&\underbrace{c_v^a(\tilde{\vu}_{12},\tilde{\vu}_3;\vlambda_{\mbox{\footnotesize vc}}^a)c_v^b(\vu_1,\vu_2;\vlambda_{\mbox{\footnotesize vc}}^b)}_{\equiv c_v(\vu_1,\vu_2,\vu_3;\vlambda_{\mbox{\footnotesize vc}})}\prod_{j=1}^3 q_j(\vtheta_j)
\end{eqnarray*}
where $\vu_j \in [0,1]^{d_j}$, $\tilde{\vu}_{12} \in [0,1]^{d_1^a}$ and $\vu_3 = \tilde{\vu}_3 \in [0,1]^{d_3}$. Here, $c_v$ is the resulting nested vector copula, $\vlambda_{\mbox{\footnotesize vc}}=(\vlambda_{\mbox{\footnotesize vc}}^{a\top},\vlambda_{\mbox{\footnotesize vc}}^{b\top})^\top$. This modular structure provides extensive flexibility, and is particularly useful as it facilitates the use of GVC-I and GVC-O (which are both limited
to two blocks) in block dependent posteriors where $M>2$. Specifically, we can choose $C_v^a$ as a GVC or KVC and $C_v^b$ as GVC-O, and use the following two-step re-parameterization trick to sample from a 3-block nested vector copula.
%, which will generate a well-defined and tractable vector copula density $c_v$ (see Appendix~\ref{nvcd} for the proof).
\begin{enumerate}
	\item Sample $(\tilde{\vu}_{12}, \tilde{\vu}_3)$ from a two-block vector copula $C_v^a$.
	\item Let $(\vu_1^\top, \vu_2^\top)^\top = \underline{\Phi}\left\{g_1(\underline{\Phi}^{-1}(\tilde{\vu}_{12}), \vlambda_{\mbox{\footnotesize vc}}^{b})\right\}$, where $g_1$ is the re-parameterization trick to sample from GVC-O.
\end{enumerate}

We next prove that $c_v(\vu_1,\vu_2,\vu_3;\vlambda_{\mbox{\footnotesize vc}})$ is a well-defined vector copula density. By definition, a vector copula density is a probability density $c(\vu_1, \ldots, \vu_M)$ defined on uniform blocks such that each variable within $\vu_j$ is distributed independently as a uniform distribution $(j = 1, \ldots, M)$. 
%In other words, the marginal densities $c_j(\vu_j) = \int c_v(\vu_1, \ldots, \vu_M) d\vu_{-j} = 1$, where $\vu_{-j} \equiv (\vu_1^\top, \ldots, \vu_{j-1}^\top, \vu_{j+1}^\top, \ldots, \vu_M^\top)^\top$.

%We show the following two-step procedure will generate a 3-block vector copula density on $\vu = (\vu_1^\top, \vu_2^\top, \vu_3^\top)$ such that 
%\begin{itemize}
%	\item Generate $(\tilde{\vu}, \vu_3)$ from a two-block vector copula $C^a$, where $\tilde{\vu} = (\tilde{\vu}_1, \tilde{\vu}_2)$.
%	\item Embed another 2-block vector copula $C^b$ in $\tilde{\vu}$. 
%\end{itemize}

%We will show the two-step re-parameterization trick will generate a nested vector copula density $c_v(\vu_1, \vu_2, \vu_3)$.
\begin{proof}
	Partition $\tilde{\vu}_{12}$ into $\tilde{\vu}_1$ and $\tilde{\vu}_2$ such that $\tilde{\vu}_{12} = (\tilde{\vu}_1^\top,\tilde{\vu}_2^\top)^\top$, where $\tilde{\vu}_1 \in [0,1]^{d_1}$ and $\tilde{\vu}_2 \in [0,1]^{d_2}$. The two step re-parameterization trick can be represented as:
	\begin{equation*}
			\begin{pmatrix}
					\vu_1 \\
					\vu_2 \\
					\vu_3
				\end{pmatrix} = \underline{\Phi}
			\begin{pmatrix}
					I_{d_1} & \vzero_{d_1 \times d_2} & \vzero_{d_1 \times d_3} \\
					Q_2 \Lambda Q_1^\top & Q_2 \sqrt{I_{\tilde{d}} - \Lambda^2} & \vzero_{d_2 \times d_3} \\
					\vzero_{d_3 \times d_1} &\vzero_{d_3 \times d_2} & I_{d_3}
				\end{pmatrix} \underline{\Phi}^{-1}
			\begin{pmatrix}
					\tilde{\vu}_1 \\
					\tilde{\vu}_2 \\
					\tilde{\vu}_3
				\end{pmatrix}
		\end{equation*}
	
	It is straightforward to see that $\vu_1 = \tilde{\vu}_1$ and $\vu_3 = \tilde{\vu}_3$. Thus, the elements within $\vu_1$ and $\vu_3$ are mutually independent, and are distributed as uniform distributions.
	
	Let $\tilde{\vz}_1 = \underline{\Phi}^{-1}(\tilde{\vu}_1)$ and  $\tilde{\vz}_2 = \underline{\Phi}^{-1}(\tilde{\vu}_2)$. Then, we have $(\tilde{\vz}_1^\top, \tilde{\vz}_2^\top)^\top \sim \mathcal{N}(0, I_{d_1+d_2})$ and $\vu_2 = \underline{\Phi}\left\{\vz_2 \right\}$ with
	\begin{equation*}
			\vz_2 =Q_2 \Lambda Q_1^\top \tilde{\vz}_1 + Q_2 \sqrt{I_{\tilde{d}} - \Lambda^2} \tilde{\vz}_2 \sim \mathcal{N}(0, I_{d_2})
		\end{equation*}
	followed by the result from Section 3.1.2 in the paper.  Thus, elements within $\vu_2$ are distributed independently as uniform distributions.
\end{proof}

\noindent \textbf{E.2: Empirical results}

\noindent We use the nested vector copula to link the global shrinkage parameter $\tilde{\xi}$ to $\valpha$ and $\tilde{\vdelta}$ in the logistic regression and correlation matrix. Specifically, we choose $C_v^b$ as GVC-I for $\valpha$ and $\tilde{\vdelta}$, and we choose $C_v^a$ as KVC. As a result, a 3-block nested vector copula is used for $\vtheta = (\valpha^\top,\tilde{\vdelta}^\top,\vxi)^\top$. We use the same marginal construction as A4, and we name this method as A7. The extended empirical results (the last row) are shown as follows.

\begin{sidewaystable}[p]
	\begin{center}
		\caption{ELBO values for different VAs (rows) in the regularized logistic regression with nine datasets (columns)}
	%	\label{tab: logitreg_ELBOs}
		\begin{tabular}{lccccccccc}
			\toprule
			\toprule
			Dataset& \multicolumn{4}{c}{Low-Dimensional Real} & \multicolumn{2}{c}{High-dimensional Real} & \multicolumn{3}{c}{High-Dimensional Simulated}\\
			\cmidrule(lr){2-5} \cmidrule(lr){6-7} \cmidrule(lr){8-10}
			Name & krkp & mushroom & spam & iono & cancer & qsar & simu1 & simu2 & simu3 \\
			Size ($n$x$m$) & 3196x38 & 8124x96 & 4601x105 & 351x112 & 42x2001 & 8992x1024 & 5Kx10K & 20Kx10K & 100Kx10K \\
			\midrule
			GMF & -382.01 & -120.85 & -856.77 & -98.82 & \textbf{-77.59} & -2245.10 & -3641.64 & -10844.22 & -40735.67 \\
			\cdashline{2-10} \\
			A4: (GVC-I \& M1-YJ) & 36.98 & \textbf{21.09} & 42.11 & 10.26 & -0.27 & \textbf{87.93} & 185.63 & 753.43 & 1929.11 \\
			A6: (GVC-I \& M2-YJ) & 33.01 & 20.05 & 40.25 & \textbf{10.55} & -1.75 & 85.45 & 121.51 & 700.10 & 1857.95 \\
			\textbf{A7: (Nested \& M1-YJ)} & \textbf{37.11} & 20.92 & \textbf{42.55} & 9.96 & -0.31 & 87.66 & \textbf{188.73} & \textbf{753.66} & \textbf{1932.40} \\
			\bottomrule
			\bottomrule
		\end{tabular}
	\end{center}
	Note: We keep the results from A4 and A6 in Table 2 because they are the optimal methods for this example. ELBO values are reported for GMF, and the differences from these values are reported for the other VAs. Higher values correspond to higher accuracy, and the bold value in each column indicates the highest ELBO value for the dataset. %The values reported are the median of ELBO values over the last 1000 steps of the SGD algorithm.
	% ``CI'' represents computationally infeasible using our Python code. Detailed descriptions of the different VAs are provided in the text.	
\end{sidewaystable}

\begin{table}[htbp]
	\begin{center}
		\caption{ELBO values for different VAs for the regularized correlation matrix $\Sigma$}
	%	\label{tab: corrmatrix_ELBOs}	
		\begin{tabular}{lccccc}
			\toprule
			\toprule
			Number of U.S. States & $r=5$ & $r=10$ &$r=20$ &$r=30$ &$r=49$ \\
			Dimension of $\vtheta$ & 21 & 91 & 381 & 871 & 2353 \\
			\midrule
			GMF & -553.93 & -1039.24 & -2038.91 & -3152.38 & -5548.02 \\
			\cdashline{2-6} \\
			A4: (GVC-I \& M1-YJ) & 10.91 & \textbf{35.48} & 71.07 & 109.90 & 181.98 \\
			A6: (GVC-I \& M2-YJ) & 10.64 & 34.96 & 70.96 & \textbf{110.06} & 181.25 \\
			\textbf{A7:  (Nested \& M1-YJ)} & \textbf{10.93} & 35.34 & \textbf{71.17} & 109.95 & \textbf{182.03} \\
			\bottomrule
			\bottomrule
		\end{tabular}
	\end{center}
	Note: We keep the results from A4 and A6 in Table 3 because they are the optimal methods for this example. The posterior is for the $(r\times r)$ regularized correlation matrix $\Sigma$ of the Gaussian copula model for U.S. wealth inequality panel data. The columns give results for $r=5$, 10, 20, 30 and 49 U.S. states, and the VAs (rows) are described
	in Section~\ref{sec:correg}. ELBO values are reported for GMF, and the differences from these values are reported for the other VAs. Higher values correspond to greater accuracy, with the largest value
	in each column in bold. 
	%The values reported are the median of ELBO values over the last 1000 steps of the SGD algorithm. 
	% DNC denotes ``Did Not Converge' within 200,000 iterations.
	%Note: The models for 5 ,10, 20, 30 and 49 states are trained by 50,50,100,150 and 200 thousand steps using Adam, respectively. For each VA, the median of ELBO values over the last 1000 steps is compared to that from GMF, and the improvement (if positive) is shown in the table. A higher value corresponds to a more accurate VA, and bold value indicates the optimal VA for a dataset. DNC denotes ``Did Not Converge' within 200 thousand iterations. We use the median here because ELBO values are highly skewed.
\end{table}

\begin{sidewaystable}[htbp]
	\begin{center}
		\caption{Computational time per 1000 steps (in proportion to GMF) for different VAs in the logistic regression}
		\begin{tabular}{lccccccccc}
			\toprule
			\toprule
			& \multicolumn{4}{c}{Low-Dimensional Real} & \multicolumn{2}{c}{High-Dimensional Real}  & \multicolumn{3}{c}{High-Dimensional Simulated} \\
			\cmidrule(lr){2-5} \cmidrule(lr){6-7} \cmidrule(lr){8-10}
			Dataset Name & krkp & mushroom & spam & iono & cancer & qsar & simu1 & simu2 & simu3 \\
			Dataset Size ($n$x$m$)& 3196x38 & 8124x96 & 4601x105 & 351x112 & 42x2001 & 8992x1024 & 5Kx10K & 20Kx10K & 100Kx10K \\
			\midrule
			GMF & 0.40 & 0.50 & 0.47 & 0.34 & 0.52 & 2.49 & 12.60 & 53.64 & 285.19 \\
			\cdashline{2-10} \\
			A4: (GVC-I \& M1+YJ) & 2.70 & 2.47 & 2.57 & 3.28 & 4.10 & 1.44 & 1.53 & 1.12 & 1.01 \\
			A6: (GVC-I \& M2+YJ) & 4.94 & 4.55 & 4.84 & 6.45 & 100.04 & 9.97 & 103.24 & 26.79 & 6.28 \\
			\textbf{A7: (Nested \& M1+YJ)} & 4.18 & 3.74 & 3.83 & 4.99 & 5.54 & 1.75 & 1.66 & 1.17 & 1.01 \\
			\bottomrule
			\bottomrule
		\end{tabular}
	\end{center}
	Note: We keep the results from A4 and A6 in Table A1 because they are the optimal methods for this example. The computational time (seconds) for GMF is reported in the table. The time of other VAs is divided by that from GMF. All VAs are trained with double precision on the CPU of Apple M3 Max. The gradients to update variational parameters are evaluated by 1000 times (100 times for simulated datasets, and then scale to 1000). The gradients are computed by Pytorch automatically. 
    %	``CI'' represents computationally infeasible.
\end{sidewaystable}

\end{document}